%% file: paper.tex
\documentclass{article}
\usepackage{iclr2017_conference,times}
\usepackage{hyperref}
\usepackage{url}
\usepackage{bm}
\usepackage{amsmath}
\usepackage{amssymb}
\usepackage{graphicx}
\usepackage{grffile}
\usepackage{pgfplots}
\usepackage{float}

\title{Lossy Image Compression with\\Compressive Autoencoders}

\author{Lucas Theis, Wenzhe Shi, Andrew Cunningham\& Ferenc Husz\'{a}r \\
Twitter \\
London, UK\\
\texttt{\{ltheis,wshi,acunningham,fhuszar\}@twitter.com}}

\iclrfinalcopy 

\begin{document}

\maketitle

\begin{abstract}
We propose a new approach to the problem of optimizing autoencoders for lossy image compression.
New media formats, changing hardware technology, as well as diverse requirements and content types
create a need for compression algorithms which are more flexible than existing codecs. Autoencoders
have the potential to address this need, but are difficult to optimize directly due to the inherent
non-differentiabilty of the compression loss. We here show that minimal changes to the
loss are sufficient to train deep autoencoders competitive with
JPEG 2000 and outperforming recently proposed approaches based on RNNs. Our network is
furthermore computationally efficient thanks to a sub-pixel architecture, which makes it suitable for high-resolution
images. This is in contrast to previous work on autoencoders for compression using coarser approximations,
shallower architectures, computationally expensive methods, or focusing on small images.
\end{abstract}

\section{Introduction}

Advances in training of neural networks have helped to improve performance in a number of
domains, but neural networks have yet to surpass existing codecs in lossy image compression.
Promising first results have recently been achieved using autoencoders
\citep{Balle:2016,Toderici:2016b} -- in particular on small images
\citep{Toderici:2016a,Gregor:2016,vanDenOord:2016b} -- and neural networks are already achieving
state-of-the-art results in lossless image compression \citep{Theis:2015,vanDenOord:2016a}.

Autoencoders have the potential to address an increasing need for flexible lossy compression algorithms.
Depending on the situation, encoders and decoders of different computational complexity are
required. When sending data from a server to a mobile device, it may be desirable to
pair a powerful encoder with a less complex decoder, but the requirements are reversed when sending data
in the other direction. The amount of computational power and bandwidth available also changes
over time as new technologies become available. For the purpose of archiving, encoding and decoding times
matter less than for streaming applications. Finally, existing compression algorithms may be far from
optimal for new media formats such as lightfield images, 360 video or VR content. While the development of a new
codec can take years, a more general compression framework based on neural networks may be able to
adapt much quicker to these changing tasks and environments.

Unfortunately, lossy compression is an inherently non-differentiable problem. In particular, quantization
is an integral part of the compression pipeline but is not differentiable. This makes it difficult
to train neural networks for this task. Existing transformations have typically
been manually chosen (e.g., the DCT transformation used in JPEG) or have been optimized for a task different from
lossy compression \citep[e.g.][used denoising autoencoders for compression]{DelTesta:2015}. In contrast to most previous work, but in line with \citet{Balle:2016}, we here aim at directly optimizing the
rate-distortion tradeoff produced by an autoencoder. We propose a simple but effective approach for
dealing with the non-differentiability of rounding-based quantization, and for approximating the
non-differentiable cost of coding the generated coefficients.

Using this approach, we achieve performance similar to or better than JPEG 2000 when evaluated for
perceptual quality. Unlike JPEG 2000, however, our framework can be optimized for
specific content (e.g., thumbnails or non-natural images), arbitrary metrics, and is readily generalizable
to other forms of media. Notably, we achieve this performance using efficient neural network architectures
which would allow near real-time decoding of large images even on low-powered consumer devices.

\section{Compressive autoencoders}
	We define a compressive autoencoder (CAE) to have three components: an encoder $f$, a decoder $g$, and a
	probabilistic model $Q$,
	\begin{align}
		f: \mathbb{R}^N \rightarrow \mathbb{R}^M, \quad
		g: \mathbb{R}^M \rightarrow \mathbb{R}^N, \quad
		Q: \mathbb{Z}^M \rightarrow [0, 1].
	\end{align}
	The discrete probability distribution defined by $Q$ is used to assign a number of bits
	to representations based on their frequencies, that is, for \textit{entropy coding}.
	All three components may have parameters and our goal is to optimize the tradeoff between using a small number of bits and
	having small distortion,
	\begin{align}
		\underbrace{-\log_2 Q\left( \left[f(\mathbf{x})\right] \right)}_\text{Number of bits}
		+ \beta \cdot \underbrace{d\left( \mathbf{x}, g( \left[f(\mathbf{x})\right] ) \right)}_\text{Distortion}.
		\label{eq:rd}
	\end{align}
	Here, $\beta$ controls the tradeoff, square brackets indicate quantization through rounding to
	the nearest integer, and $d$ measures the distortion introduced by coding and decoding. The
	quantized output of the encoder is the code used to represent an image and is stored losslessly.
	The main source of information loss is the quantization (Appendix~\ref{sec:noround}). Additional information may be discarded
	by the encoder, and the decoder may not perfectly decode the available information, increasing
	distortion.
	
	Unfortunately we cannot optimize Equation~\ref{eq:rd} directly using gradient-based techniques, as
	$Q$ and $\left[\cdot\right]$ are non-differentiable. The following two sections propose a solution
	to deal with this problem.

	\subsection{Quantization and differentiable alternatives}
		\label{sec:quantization}
		\begin{figure}[t]
			\centering
			\input{figures/quantization.tex}
			\vspace{-.5cm}
			\caption{Effects of rounding and differentiable alternatives when used as replacements in JPEG compression.
			\textbf{A}: A crop of an image before compression \citep{GoToVan:2014}. \textbf{B}: Blocking artefacts in JPEG are caused by rounding
			of DCT coefficients to the nearest integer. Since rounding is used at test time, a good
			approximation should produce similar artefacts. \textbf{C}: Stochastic rounding to the nearest
			integer similar to the binarization of \citet{Toderici:2016a}. \textbf{D}: Uniform additive noise
			\citep{Balle:2016}.}
			\label{fig:quantization}
		\end{figure}
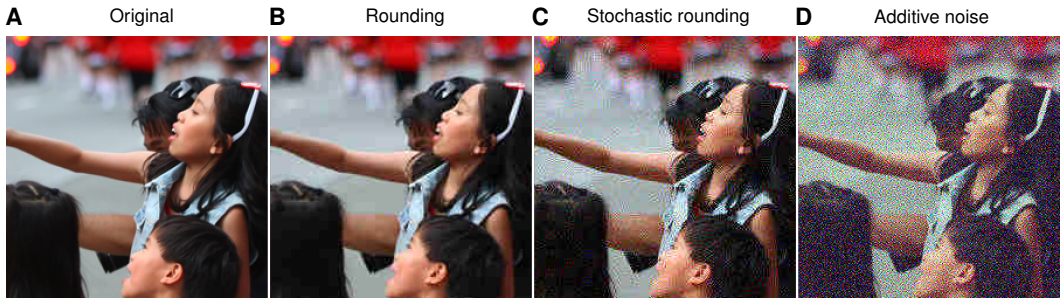

		The derivative of the rounding function is zero everywhere except at integers, where it is
		undefined. We propose to replace its derivative in the backward pass of backpropagation
		\citep{Rumelhart:1986} with the derivative of a smooth approximation, $r$, that
		is, effectively defining the derivative to be
		\begin{align}
			\frac{d}{dy} \left[ y \right] := \frac{d}{dy} r(y).
		\end{align}
		Importantly, we do not fully replace the rounding function with a smooth approximation but only
		its derivative, which means that quantization is still performed as usual in the forward pass. If we replaced rounding with
		a smooth approximation completely, the decoder might learn to invert the smooth
		approximation, thereby removing the information bottle neck that forces the network to
		compress information.

		Empirically, we found the identity, $r(y) = y$, to work as well as more sophisticated choices.
		This makes this operation easy to implement, as we simply have to pass gradients without
		modification from the decoder to the encoder.

		Note that the gradient with respect to the decoder's parameters can be computed without resorting to
		approximations, assuming $d$ is differentiable. In contrast to related approaches, our approach
		has the advantage that it does not change the gradients of the decoder, since the forward pass
		is kept the same.

		In the following, we discuss alternative approaches proposed by other authors. Motivated by theoretical links to dithering,
		\citet{Balle:2016} proposed to replace quantization by additive uniform noise,
		\begin{align}
			\left[ f(\mathbf{x}) \right] \approx f(\mathbf{x}) + \mathbf{u}.
		\end{align}
		\citet{Toderici:2016a}, on the other hand, used a stochastic form of binarization
		\citep{Williams:1992}. Generalizing this idea to integers, we define the following stochastic
		rounding operation:
		\begin{align}
			\left\{ y \right\} \approx \lfloor y \rfloor + \bm{\varepsilon}, \quad
			\bm{\varepsilon} \in \{ 0, 1 \}, \quad
			P(\bm{\varepsilon} = 1) = y - \lfloor y \rfloor,
		\end{align}
		where $\lfloor \cdot \rfloor$ is the floor operator. In the backward pass, the derivative is
		replaced with the derivative of the expectation,
		\begin{align}
			\frac{d}{dy}\left\{ y \right\} := \frac{d}{dy}\mathbb{E}\left[ \left\{ y \right\} \right] = \frac{d}{dy} y = 1.
		\end{align}
		Figure~\ref{fig:quantization} shows the effect of using these two alternatives
		as part of JPEG, whose encoder and decoder are based on a block-wise DCT transformation
		\citep{Pennebaker:1993}. Note that the output is visibly different from the output produced
		with regular quantization by rounding and that the error signal sent to the autoencoder depends on these images.
		Whereas in Fig.~\ref{fig:quantization}B the error signal received by the decoder would be to remove blocking
		artefacts, the signal in Fig.~\ref{fig:quantization}D will be to remove high-frequency
		noise. We expect this difference to be less of a problem with simple metrics such as mean-squared error
		and to have a bigger impact when using more perceptually meaningful measures of distortion.

		An alternative would be to use the latter approximations only for the gradient of the encoder but
		not for the gradients of the decoder. While this is possible, it comes at the cost of increased
		computational and implementational complexity, since we would have to perform the forward
		and backward pass through the decoder twice: once using rounding, once using the
		approximation. With our approach the gradient of the decoder is correct even for a
		single forward and backward pass.

	\subsection{Entropy rate estimation}
		Since $Q$ is a discrete function, we cannot differentiate it with respect to its argument,
		which prevents us from computing a gradient for the encoder. To solve this problem, we use a
		continuous, differentiable approximation. We upper-bound the non-differentiable number of
		bits by first expressing the model's distribution $Q$ in terms of a probability density $q$,
		\begin{align}
			Q(\mathbf{z}) = \int_{[-.5, .5[^M} q(\mathbf{z} + \mathbf{u}) \, d\mathbf{u}.
		\end{align}
		An upper bound is given by:
		\begin{align}
			-\log_2 Q\left( \mathbf{z} \right)
			= -\log_2 \int_{[-.5, .5[^M} q(\mathbf{z} + \mathbf{u}) \, d\mathbf{u}
			\leq \int_{[-.5, .5[^M} -\log_2 q(\mathbf{z} + \mathbf{u}) \, d\mathbf{u},
		\end{align}
		where the second step follows from Jensen's inequality \citep[see also][]{Theis:2016}.
		An unbiased estimate of the upper bound is obtained by sampling $\mathbf{u}$ from the unit cube
		$[-.5, .5[^M$. If we use a differentiable density, this estimate will be differentiable in
		$\mathbf{z}$ and therefore can be used to train the encoder.

	\subsection{Variable bit rates}
		\label{sec:variable}
		In practice we often want fine-gained control over the number of bits used. One way to achieve this
		is to train an autoencoder for different rate-distortion tradeoffs. But this would require
		us to train and store a potentially large number of models. To reduce these costs, we
		finetune a pre-trained autoencoder for different rates by introducing scale
		parameters\footnote{To ensure positivity, we use a different parametrization and optimize log-scales rather than
		scales.} $\bm{\lambda} \in \mathbb{R}^M$,
		\begin{align}
			\label{eq:scales}
			-\log_2 q\left( \left[f(\mathbf{x}) \circ \bm{\lambda} \right] + \mathbf{u} \right)
			+ \beta \cdot d\left( \mathbf{x}, g( \left[f(\mathbf{x}) \circ \bm{\lambda} \right] / \bm{\lambda} ) \right).
		\end{align}
		Here, $\circ$ indicates point-wise multiplication and division is also performed point-wise.
		To reduce the number of trainable scales, they may furthermore be shared across dimensions.
		Where $f$ and $g$ are convolutional, for example, we share scale parameters across spatial
		dimensions but not across channels.

		An example of learned scale parameters is shown in Figure~\ref{fig:training}A.
		For more fine-grained control over bit rates, the optimized scales can be interpolated.

	\subsection{Related work}
		\label{sec:related}
		Perhaps most closely related to our work is the work of \citet{Balle:2016}. The main differences
		lie in the way we deal with quantization (see Section~\ref{sec:quantization}) and
		entropy rate estimation. The transformations used by \citet{Balle:2016} consist of a single
		linear layer combined with a form of contrast gain control, while our framework relies on
		more standard deep convolutional neural networks. 

		\citet{Toderici:2016a} proposed to use recurrent neural networks (RNNs) for compression.
		Instead of entropy coding as in our work, the network tries to minimize the distortion for a
		given number of bits. The image is encoded in an iterative manner, and decoding is performed in each
		step to be able to take into account residuals at the next iteration.
		An advantage of this design is that it allows for progressive coding of images. A disadvantage is that compression
		is much more time consuming than in our approach, as we use efficient convolutional neural networks and
		do not necessarily require decoding at the encoding stage.

		\citet{Gregor:2016} explored using variational autoencoders with recurrent encoders and decoders for
		compression of small images. This type of autoencoder is trained to maximize the lower bound of a log-likelihood,
		or equivalently to minimize
		\begin{align}
			-\mathbb{E}_{p(\mathbf{y} \mid \mathbf{x})}\left[ \log \frac{q(\mathbf{y}) q(\mathbf{x} \mid \mathbf{y})}{p(\mathbf{y} \mid \mathbf{x})} \right],
		\end{align}
		where $p(\mathbf{y} \mid \mathbf{x})$ plays the role of the encoder, and $q(\mathbf{x} \mid
		\mathbf{y})$ plays the role of the decoder. While \citet{Gregor:2016} used a Gaussian
		distribution for the encoder, we can link their approach to the work of \cite{Balle:2016} by
		assuming it to be uniform, $p(\mathbf{y} \mid \mathbf{x}) = f(\mathbf{x}) + \mathbf{u}$. If we also assume a Gaussian
		likelihood with fixed variance, $q(\mathbf{x} \mid \mathbf{y}) = \mathcal{N}(\mathbf{x} \mid g(\mathbf{y}), \sigma^2
		\mathbf{I})$, the objective function can be written
		\begin{align}
			\mathbb{E}_{\mathbf{u}}\left[ -\log q(f(\mathbf{x}) + \mathbf{u}) + \frac{1}{2\sigma^2} || \mathbf{x} - g(f(\mathbf{x}) + \mathbf{u})||^2 \right] + C.
		\end{align}
		Here, $C$ is a constant which encompasses the negative entropy of the encoder and the
		normalization constant of the Gaussian likelihood. Note that this equation is identical to a
		rate-distortion trade-off with $\beta = \sigma^{-2}/2$ and quantization replaced by
		additive uniform noise. However, not all distortions have an equivalent formulation as a
		variational autoencoder \citep{Kingma:2014}. This only works if $e^{-d(\mathbf{x}, \mathbf{y})}$ is normalizable
		in $\mathbf{x}$ and the normalization constant does not depend on $\mathbf{y}$, or otherwise
		$C$ will not be constant. An direct empirical comparison of our approach with variational autoencoders is
		provided in Appendix~\ref{sec:vae}.

		\citet{Ollivier:2015} discusses variational autoencoders for lossless compression as well as
		connections to denoising autoencoders.


\section{Experiments}
	\subsection{Encoder, decoder, and entropy model}
		\begin{figure}[t]
			\centering
			\hbox{
				\hspace{-1cm}
				\includegraphics[width=1.15\textwidth]{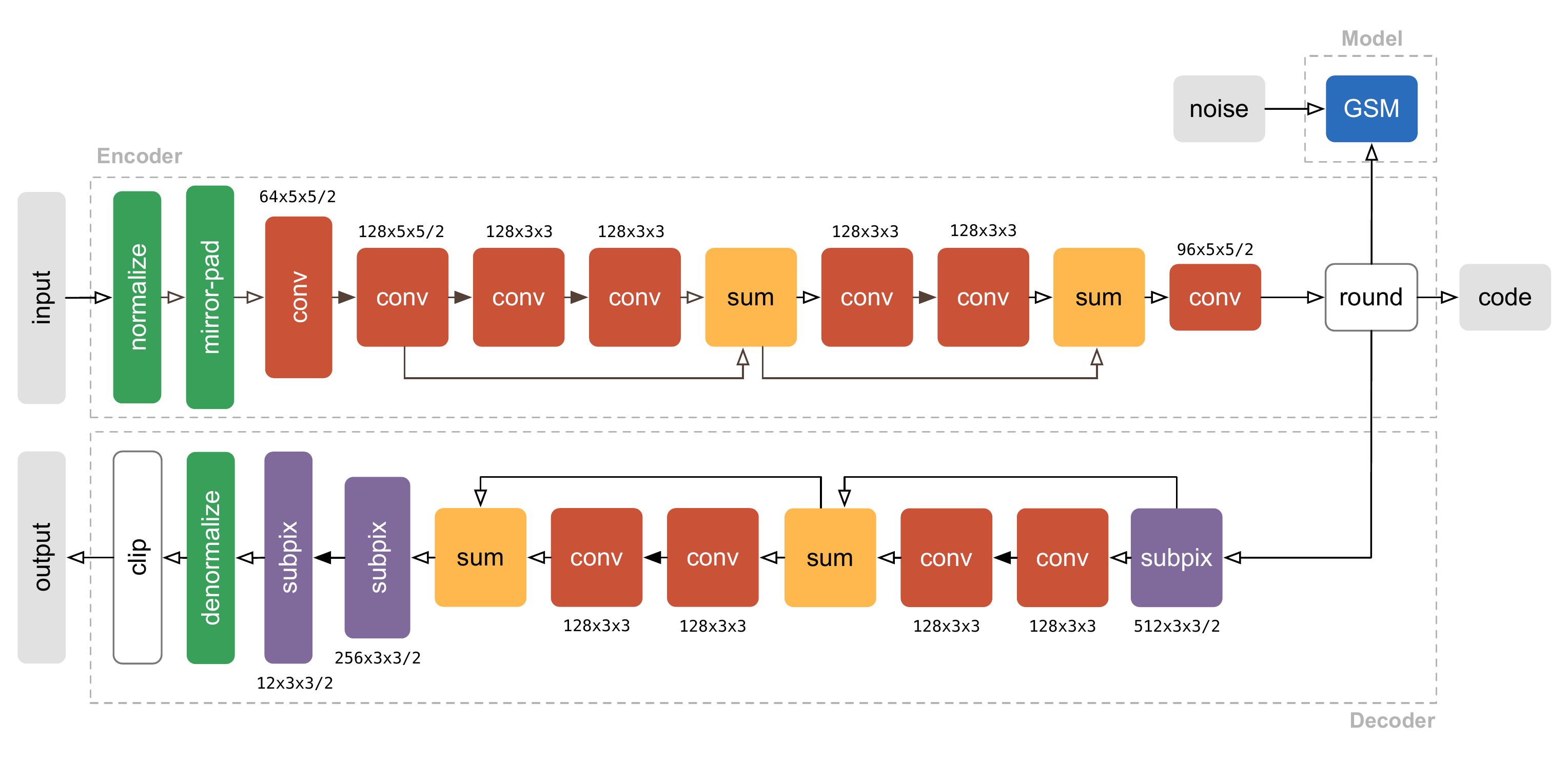}
			}
			\caption{Illustration of the compressive autoencoder architecture used in this paper. Inspired by the work
			of \citet{Shi:2016}, most convolutions are performed in a downsampled space to speed up
			computation, and upsampling is performed using sub-pixel convolutions (convolutions followed by
			reshaping/reshuffling of the coefficients). To reduce clutter, only two residual blocks
			of the encoder and the decoder are shown. Convolutions followed by leaky rectifications
			are indicated by solid arrows, while transparent arrows indicate absence of additional
			nonlinearities. As a model for the distributions of quantized coefficients we use Gaussian scale
			mixtures. The notation $C \times K \times K$ refers to $K \times K$ convolutions with $C$ filters. The number
			following the slash indicates stride in the case of convolutions, and upsampling factors
			in the case of sub-pixel convolutions.}
			\label{fig:cae}
		\end{figure}

		We use common convolutional neural networks \citep{LeCun:1998} for the encoder and the decoder of the compressive
		autoencoder. Our architecture was inspired by the work of \citet{Shi:2016}, who
		demonstrated that super-resolution can be achieved much more efficiently by operating in the
		low-resolution space, that is, by convolving images and then upsampling instead of upsampling
		first and then convolving an image.

		The first two layers of the encoder perform preprocessing, namely mirror padding and
		a fixed pixel-wise normalization. The mirror-padding was chosen such that the output of
		the encoder has the same spatial extent as an 8 times downsampled image. The normalization
		centers the distribution of each channel's values and ensures it has approximately unit
		variance. Afterwards, the image is convolved and spatially downsampled while at the same time increasing the number of channels to 128. This is followed by
		three residual blocks \citep{He:2015}, where each block consists of an additional two
		convolutional layers with 128 filters each. A final convolutional layer is applied
		and the coefficients downsampled again before quantization through rounding to the nearest
		integer.

		The decoder mirrors the architecture of the encoder (Figure~\ref{fig:cae}). Instead of
		mirror-padding and \textit{valid} convolutions, we use zero-padded convolutions.
		Upsampling is achieved through convolution followed by a reorganization of the coefficients. This reorganization turns a tensor with many channels into a
		tensor of the same dimensionality but with fewer channels and larger spatial extent
		\citep[for details, see][]{Shi:2016}. A convolution and reorganization of coefficients
		together form a \textit{sub-pixel convolution layer}. Following three residual blocks,
		two sub-pixel convolution layers upsample the image to the resolution of the
		input. Finally, after denormalization, the pixel values are clipped to the range of 0 to 255.
		Similar to how we deal with gradients of the rounding function, we redefine the gradient
		of the clipping function to be 1 outside the clipped range. This ensures that the training signal is
		non-zero even when the decoded pixels are outside this range (Appendix~\ref{sec:clipping}).

		To model the distribution of coefficients and estimate the bit rate, we use independent Gaussian scale mixtures (GSMs),
		\begin{align}
			\log_2 q(\mathbf{z} + \mathbf{u}) = \sum_{i,j,k} \log_2 \sum_s \pi_{ks} \mathcal{N}(z_{kij} + u_{kij}; 0, \sigma_{ks}^2),
		\end{align}
		where $i$ and $j$ iterate over spatial positions, and $k$ iterates over channels of the
		coefficients for a single image $\mathbf{z}$. GSMs are well established as useful building
		blocks for modelling filter responses of natural images \citep[e.g.,][]{Portilla:2003}. We
		used 6 scales in each GSM.
		Rather than using the more common parametrization above, we parametrized the GSM so that it can be
		easily used with gradient based methods, optimizing log-weights and log-precisions rather
		than weights and variances. We note that the
		leptokurtic nature of GSMs \citep{Andrews:1974} means that the rate term encourages sparsity
		of coefficients.

		All networks were implemented in Python using Theano \citeyearpar{Theano:2016} and Lasagne
		\citep{Dieleman:2015}. For entropy encoding of the quantized coefficients, we first created
		Laplace-smoothed histogram estimates of the coefficient distributions across a training set.
		The estimated probabilities were then used with a publicly available BSD licensed implementation of a range
		coder\footnote{\url{https://github.com/kazuho/rangecoder/}}.

	\subsection{Incremental training}
		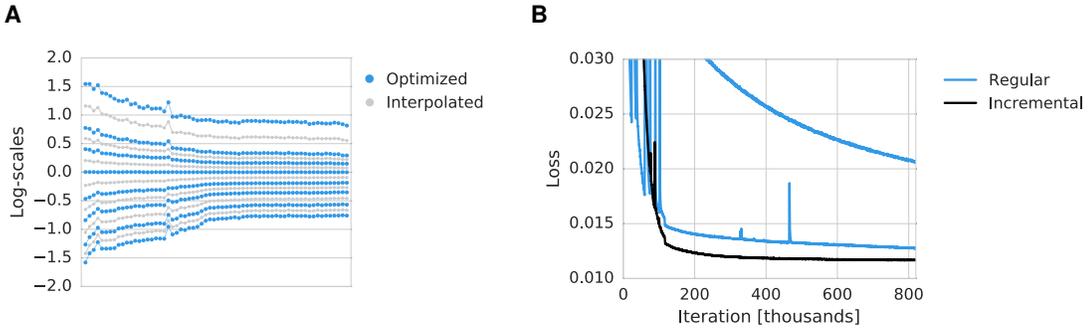
\begin{figure}[t]
			\centering
			\input{figures/training.tex}
			\vspace{-.7cm}
			\caption{\textbf{A}: Scale parameters obtained by finetuning a compressive autoencoder
			(blue). More fine-grained control over bit rates can be achieved by interpolating
			scales (gray). Each dot corresponds to the scale parameter of one coefficient for a
			particular rate-distortion trade-off. The coefficients are ordered due to the incremental training procedure.
			\textbf{B}: Comparison of incremental training versus non-incremental
			training. The learning rate was decreased after 116,000 iterations (bottom two lines).
			Non-incremental training is initially less stable and shows worse performance at later iterations.
			Using a small learning rate from the beginning stabilizes non-incremental training but is considerably slower (top line).}
			\label{fig:training}
		\end{figure}

		All models were trained using \textit{Adam} \citep{Kingma:2015} applied to batches of 32 images $128 \times 128$ pixels in size.
		We found it beneficial to optimize coefficients in an incremental manner
		(Figure~\ref{fig:training}B). This is done by introducing an additional
		binary mask $\mathbf{m}$,
		\begin{align}
			\label{eq:masks}
			-\log_2 q\left( \left[f(\mathbf{x}) \right] \circ \mathbf{m} + \mathbf{u} \right)
			+ \beta \cdot d\left( \mathbf{x}, g( \left[f(\mathbf{x}) \right] \circ \mathbf{m} ) \right).
		\end{align}
		Initially, all but 2 entries of the mask are set to zero. Networks are
		trained until performance improvements reach below a threshold, and then another
		coefficient is enabled by setting an entry of the binary mask to 1.
		After all coefficients have been enabled, the learning rate is reduced from an initial value of $10^{-4}$ to
		$10^{-5}$. Training was performed for up to $10^6$ updates but usually reached good
		performance much earlier.

		After a model has been trained for a fixed rate-distortion trade-off ($\beta$), we introduce and
		fine-tune scale parameters (Equation~\ref{eq:scales}) for other values of $\beta$ while keeping
		all other parameters fixed. Here we used an initial learning rate of $10^{-3}$ and
		continuously decreased it by a factor of $\tau^\kappa / (\tau + t)^\kappa$, where $t$ is the
		current number of updates performed, $\kappa = .8$, and $\tau = 1000$. Scales were optimized for 10,000
		iterations. For even more fine-grained control over the bit rates, we interpolated between
		scales optimized for nearby rate-distortion tradeoffs.

	\subsection{Natural images}
		\begin{figure}[t]
			\centering
			\hbox {
				\hspace{-1cm}
				\input{figures/quantitative_comparison.tex}
			}
			\caption{Comparison of different compression algorithms with respect to PSNR, SSIM, and
			MS-SSIM on the Kodak PhotoCD image dataset. We note that the blue line refers to the results of
			\citet{Toderici:2016b} achieved \textit{without} entropy encoding.}
			\label{fig:comparison}
		\end{figure}
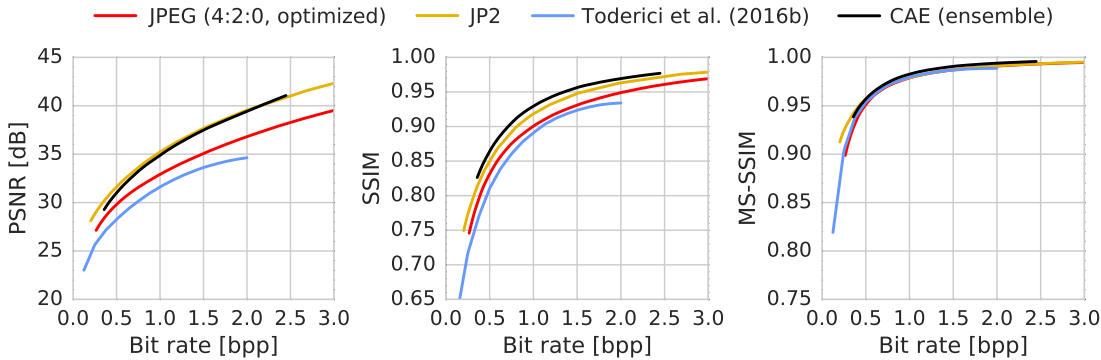

		We trained compressive autoencoders on 434 high quality images licensed under creative commons and
		obtained from \href{http://flickr.com}{flickr.com}. The images were downsampled to below $1536 \times 1536$
		pixels and stored as lossless PNGs to avoid compression artefacts. From these images, we extracted
		$128 \times 128$ crops to train the network. Mean squared error was used as a measure of distortion
		during training. Hyperparameters affecting network architecture and training were evaluated on a
		small set of held-out Flickr images. For testing, we use the commonly used Kodak PhotoCD
		dataset of 24 uncompressed $768 \times 512$ pixel images\footnote{\url{http://r0k.us/graphics/kodak/}}.

		We compared our method to JPEG \citep{Wallace:1991}, JPEG 2000 \citep{Skodras:2001}, and the RNN-based method of
		\citep{Toderici:2016b}\footnote{We used the code which was made available on
		\url{https://github.com/tensorflow/models/tree/2390974a/compression}. We note that at the time of
		this writing, this implementation does not include entropy coding as in the paper of \citet{Toderici:2016b}.}.
		Bits for header information were not counted towards the bit rate of JPEG and JPEG 2000.
		Among the different variants of JPEG, we found that optimized JPEG with 4:2:0 chroma
		sub-sampling generally worked best (Appendix~\ref{sec:jpeg}).

		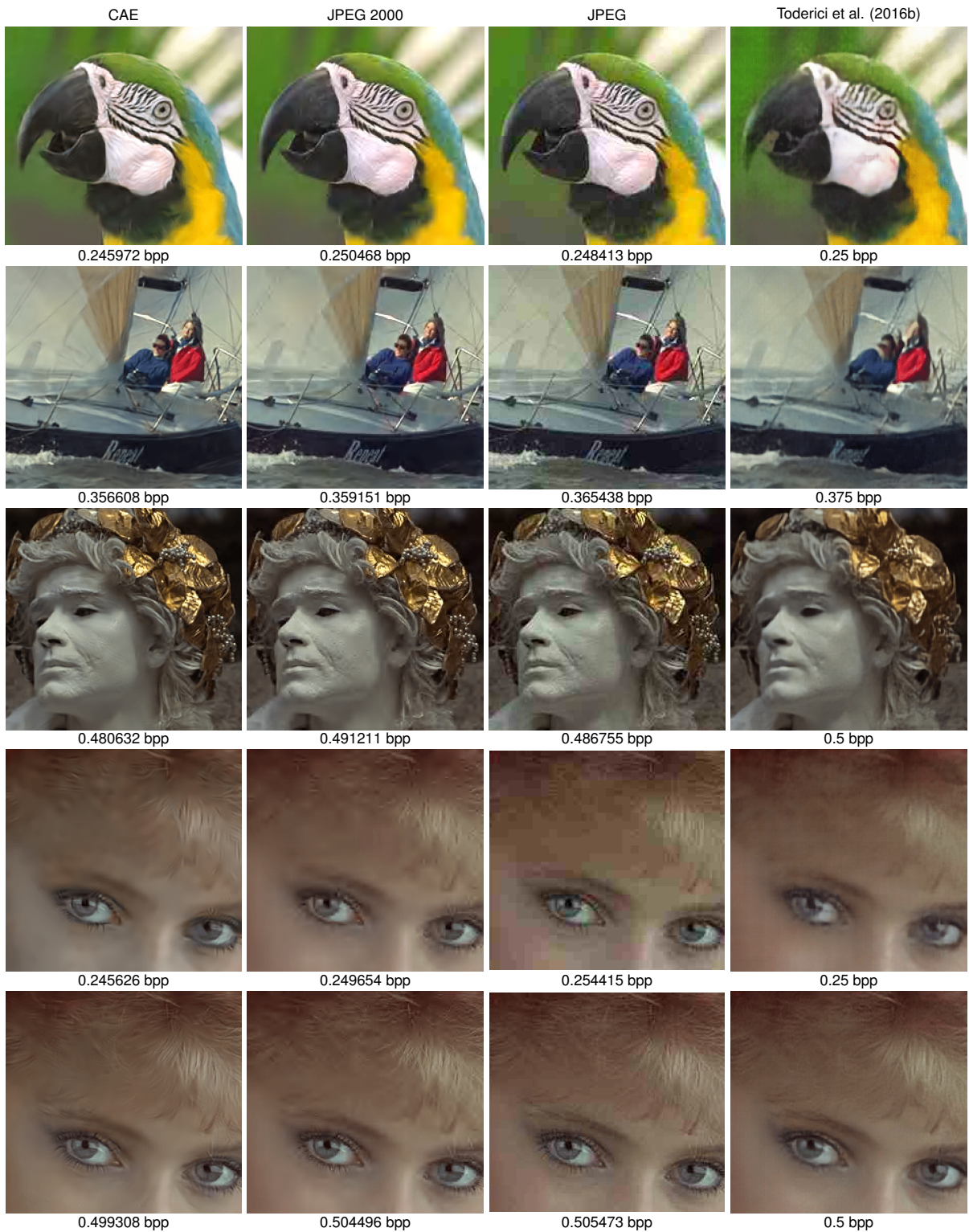
\begin{figure}[p]
			\centering
			\hbox {
				\hspace{-1cm}
				\input{figures/closeups.tex}
			}
			\caption{Closeups of images produced by different compression algorithms at relatively low
			bit rates. The second row shows an example where our method performs well, producing
			sharper lines than and fewer artefacts than other methods. The fourth row shows an example where
			our method struggles, producing noticeable artefacts in the hair and discolouring the skin. At higher bit rates,
			these problems disappear and CAE reconstructions appear sharper than those of
			JPEG 2000 (fifth row). Complete images are provided in Appendix~\ref{sec:details}.}
			\label{fig:closeups}
		\end{figure}

		While fine-tuning a single compressive autoencoder for a wide range of bit rates
		worked well, optimizing all parameters of a network for a particular rate distortion
		trade-off still worked better. We here chose the compromise of combining autoencoders trained
		for low, medium or high bit rates (see Appendix~\ref{sec:ensemble} for details).

		For each image and bit rate, we choose the autoencoder producing the smallest distortion.
		This increases the time needed to compress an image, since an image has to be encoded and decoded multiple
		times. However, decoding an image is still as fast, since it only requires choosing and running one
		decoder network. A more efficient but potentially less performant solution would be to always choose the same autoencoder
		for a given rate-distortion tradeoff. We added 1 byte to the coding cost to encode which autoencoder of an ensemble
		is used.

		Rate-distortion curves averaged over all test images are shown in Figure~\ref{fig:comparison}. We
		evaluated the different methods in terms of PSNR, SSIM \citep{Wang:2004a}, and multiscale SSIM
		\citep[MS-SSIM; ][]{Wang:2004b}. We used the implementation of \citet{scikit-image:2014} for
		SSIM and the implementation of \citet{Toderici:2016b} for MS-SSIM. We find that in terms of PSNR,
		our method performs similar to JPEG 2000 although slightly worse at low and medium bit rates and slightly better at
		high bit rates. In terms of SSIM, our method outperforms all other tested methods. MS-SSIM produces
		very similar scores for all methods, except at very low bit rates. However, we also find
		these results to be highly image dependent. Results for individual images are provided as
		supplementary material\footnote{\url{https://figshare.com/articles/supplementary_zip/4210152}}.

		In Figure~\ref{fig:closeups} we show crops of images compressed to low bit rates.
		In line with quantitative results, we find that JPEG 2000 reconstructions appear visually
		more similar to CAE reconstructions than those of other methods. However, artefacts produced
		by JPEG 2000 seem more noisy than CAE's, which are smoother and sometimes appear G\'{a}bor-filter-like.

		To quantify the subjective quality of compressed images, we ran a \textit{mean opinion
		score} (MOS) test. While MOS tests have their limitations, they are a widely used
		standard for evaluating perceptual quality \citep{Streijl:2014}. Our MOS test set included the 24 full-resolution uncompressed originals
		from the Kodak dataset, as well as the same images compressed using each of four algorithms
		at or near three different bit rates: $0.25$, $0.372$ and $0.5$ bits per pixel. Only the
		low-bit-rate CAE was included in this test.

		For each image, we chose the CAE setting which produced the highest bit rate but did not exceed the
		target bit rate. The average bit rates of CAE compressed images were $0.24479$, $0.36446$, and
		$0.48596$, respectively. We then chose the smallest quality factor for JPEG and JPEG 2000 for
		which the bit rate exceeded that of the CAE. The average bit rates for JPEG were $0.25221$,
		$0.37339$ and $0.49534$, for JPEG 2000 $0.24631$, $0.36748$ and $0.49373$. For some images
		the bit rate of the CAE at the lowest setting was still higher than the target bit rate.
		These images were excluded from the final results, leaving 15, 21, and 23 images,
		respectively.

		The perceptual quality of the resulting $273$ images was rated by $n=24$ non-expert evaluators.
		One evaluator did not finish the experiment, so her data was discarded. The images were presented to
		each individual in a random order. The evaluators gave a discrete opinion score for each image
		from a scale between $1$ (bad) to $5$ (excellent). Before the rating began, subjects were presented
		an uncompressed calibration image of the same dimensions as the test images (but not from the Kodak
		dataset). They were then shown four versions of the calibration image using the worst quality setting
		of all four compression methods, and given the instruction ``These are examples of compressed images.
		These are some of the worst quality examples.''

\begin{figure}[t]
			\begin{center}
				\includegraphics[width=3in]{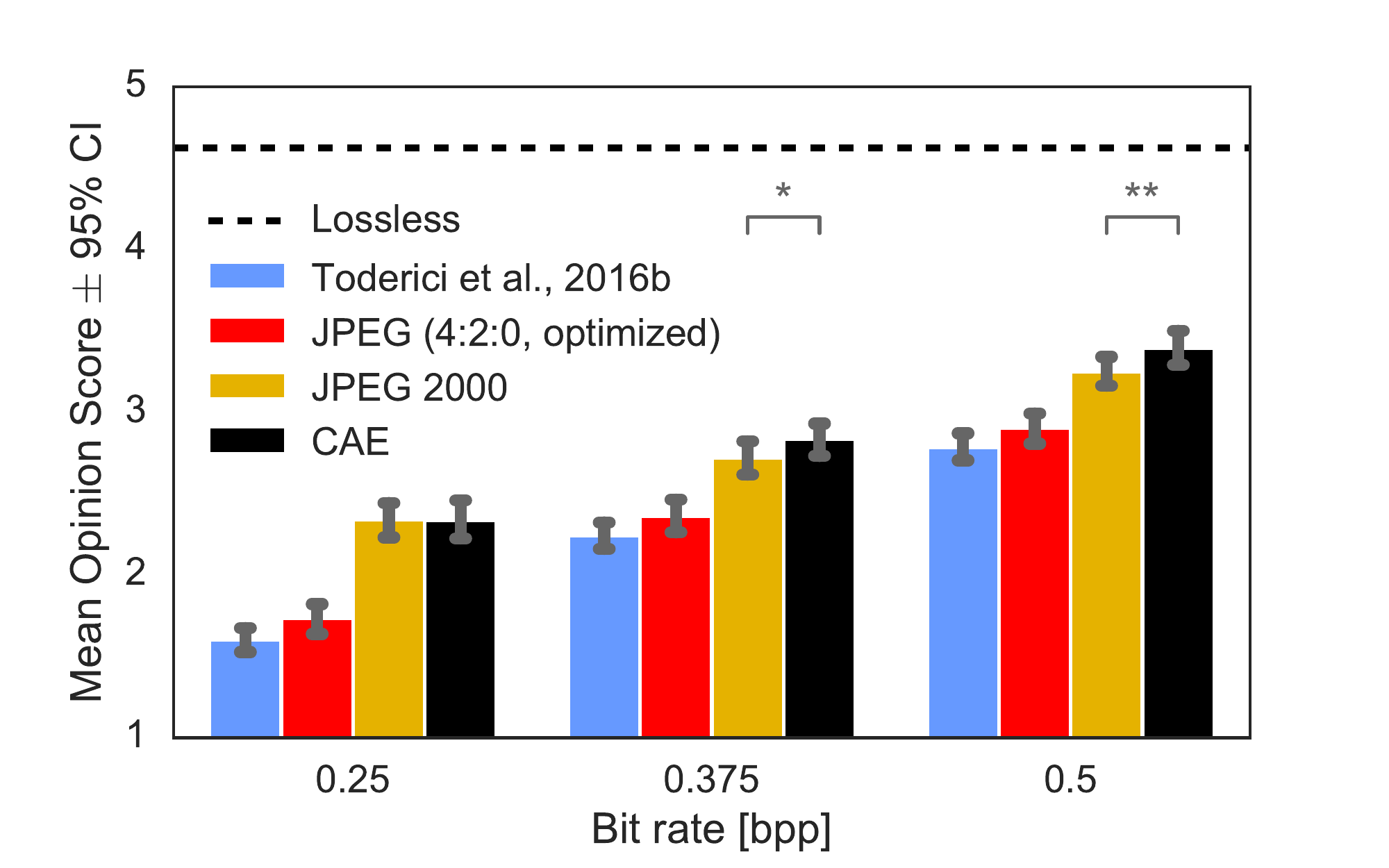}
			\end{center}
			\vspace{-.5cm}
			\caption{Results of a mean opinion score test.}
			\label{fig:mos}
		\end{figure}

		Figure~\ref{fig:mos} shows average MOS results for each algorithm at each bit rate.
		95\% confidence intervals were computed via bootstrapping. We found that CAE and JPEG 2000 achieved higher MOS
		than JPEG or the method of \citet{Toderici:2016b} at all bit rates we tested.
		We also found that CAE significantly outperformed JPEG 2000 at $0.375\, bpp$ ($p<0.05$) and
		$0.5\, bpp$ ($p<0.001$).

\section{Discussion}
	We have introduced a simple but effective way of dealing with non-differentiability in
	training autoencoders for lossy compression. Together with an incremental training strategy, this
	enabled us to achieve better performance than JPEG 2000 in terms of SSIM and MOS scores.
	Notably, this performance was achieved using an efficient convolutional architecture,
	combined with simple rounding-based quantization and a simple entropy coding scheme.
	Existing codecs often benefit from hardware support, allowing them to run at low energy costs. However,
	hardware chips optimized for convolutional neural networks are likely to be widely available soon,
	given that these networks are now key to good performance in so many applications.

	While other trained algorithms have been shown to provide similar results as JPEG 2000 \citep[e.g.][]{vanDenOord:2014},
	to our knowledge this is the first time that an end-to-end trained architecture has been demonstrated
	to achieve this level of performance on high-resolution images. An \textit{end-to-end} trained autoencoder has the
	advantage that it can be optimized for arbitrary metrics. Unfortunately, research on perceptually relevant metrics suitable for optimization is
	still in its infancy \citep[e.g.,][]{Dosovitskiy:2016,Balle:2016}. While perceptual metrics exist which
	correlate well with human perception for certain types of distortions
	\citep[e.g.,][]{Wang:2004a,Laparra:2016}, developing a perceptual metric which can be optimized is a more
	challenging task, since this requires the metric to behave well for a much larger variety of
	distortions and image pairs.

	In future work, we would like to explore the optimization of compressive autoencoders for
	different metrics. A promising direction was presented by \citet{Bruna:2016}, who achieved interesting super-resolution
	results using metrics based on neural networks trained for image classification. \citet{Gatys:2016}
	used similar representations to achieve a breakthrough in perceptually meaningful style transfer.
	An alternative to perceptual metrics may be to use generative adversarial networks
	\citep[GANs;][]{Goodfellow:2014}. Building on the work of \citet{Bruna:2016} and
	\citet{Dosovitskiy:2016}, \citet{Ledig:2016} recently demonstrated impressive super-resolution results
	by combining GANs with feature-based metrics.

\subsubsection*{Acknowledgments}
	We would like to thank Zehan Wang, Aly Tejani, Cl\'{e}ment Farabet, and Luke Alonso for helpful feedback on the manuscript.

\bibliography{references}
\bibliographystyle{references}

\appendix
\section{Appendix}

	\subsection{Gradient of clipping}
		\label{sec:clipping}
		We redefine the gradient of the clipping operation to be constant,
		\begin{align}
			\frac{d}{d\hat x} \text{clip}_{0,255}(\hat x) := 1.
		\end{align}
		Consider how this affects the gradients of a squared loss,
		\begin{align}
			\frac{d}{d\hat x} (\text{clip}_{0,255}(\hat x) - x)^2
			= 2 (\text{clip}_{0,255}(\hat x) - x) \frac{d}{d\hat x} \text{clip}_{0,255}(\hat x).
		\end{align}
		Assume that $\hat x$ is larger than 255. Without redefinition of the
		derivative, the error signal will be 0 and not helpful. Without any
		clipping, the error signal will depend on the value of $\hat x$, even though any value above
		255 will have the same effect on the loss at test time. On the other hand, using clipping
		but a different signal in the backward pass is intuitive, as it yields an error signal which
		is proportional to the error that would also be incurred at test time.

	\subsection{Different modes of JPEG}
		\label{sec:jpeg}

		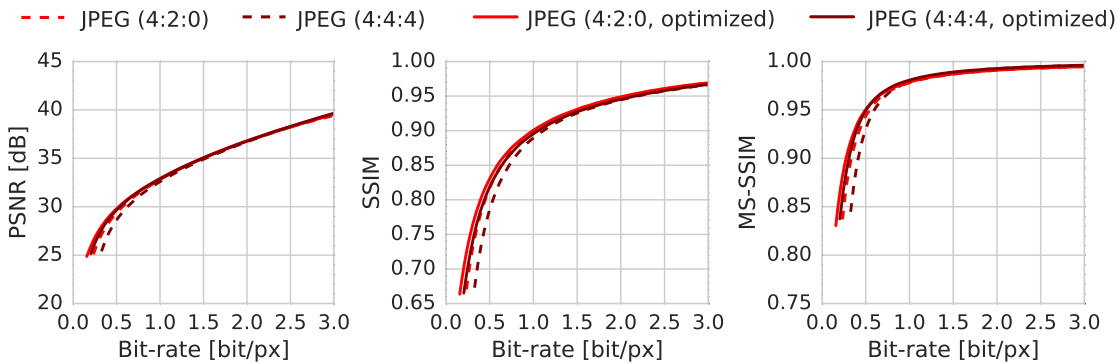
\begin{figure}[h!]
			\centering
			\hbox{
				\hspace{-1cm}
				\input{figures/jpeg.tex}
			}
			\caption{A comparison of different JPEG modes on the Kodak PhotoCD image dataset.
			Optimized Huffman tables perform better than default Huffman tables.}
			\label{fig:jpeg}
		\end{figure}

		We compared optimized and non-optimized JPEG with (4:2:0) and without (4:4:4) chroma sub-sampling.
		Optimized JPEG computes a Huffman table specific to a given image, while unoptimized JPEG uses a
		predefined Huffman table. We did not count bits allocated to the header of the file format, but for optimized
		JPEG we counted the bits required to store the Huffman table. We found that on average,
		chroma-subsampled and optimized JPEG performed better on the Kodak dataset (Figure~\ref{fig:jpeg}).

	\subsection{Compression vs dimensionality reduction}
		\label{sec:noround}
		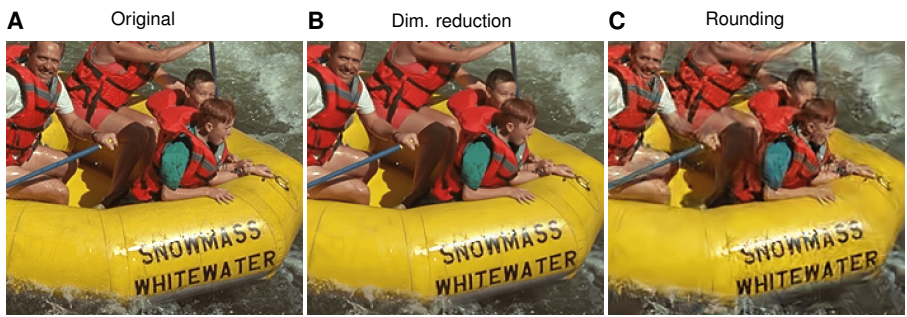
\begin{figure}[h!]
			\centering
			\input{figures/round_vs_noround.tex}
			\caption{To disentangle the effects of quantization and dimensionality reduction, we
			reconstructed images with quantization disabled. \textbf{A}: The original uncompressed
			image. \textbf{B}: A reconstruction generated by a compressive autoencoder, but with the
			rounding operation removed. The dimensionality of the encoder's output is $3\times$
			smaller than the input. \textbf{C}: A reconstruction generated by the same compressive
			autoencoder. While the effects of dimensionality reduction are almost imperceptible, quantization
			introduces visible artefacts.}
			\label{fig:noround}
		\end{figure}

		Since a single real number can carry as much information as a high-dimensional entity, dimensionality
		reduction alone does not amount to compression. However, if we constrain the architecture of the encoder,
		it may be forced to discard certain information. To better understand how much information
		is lost due to dimensionality reduction and how much information is lost due to
		quantization, Figure~\ref{fig:noround} shows reconstructions produced by a compressive
		autoencoder with and without quantization. The effect of dimensionality reduction is minimal
		compared to the effect of quantization.

	\subsection{Ensemble}
		\label{sec:ensemble}
		\begin{figure}[H]
			\centering
			\hbox {
				\hspace{-1cm}
				\input{figures/quantitative_comparison_cae.tex}
			}
			\caption{Comparison of CAEs optimized for low, medium, or high bit rates.}
			\label{fig:cae}
		\end{figure}
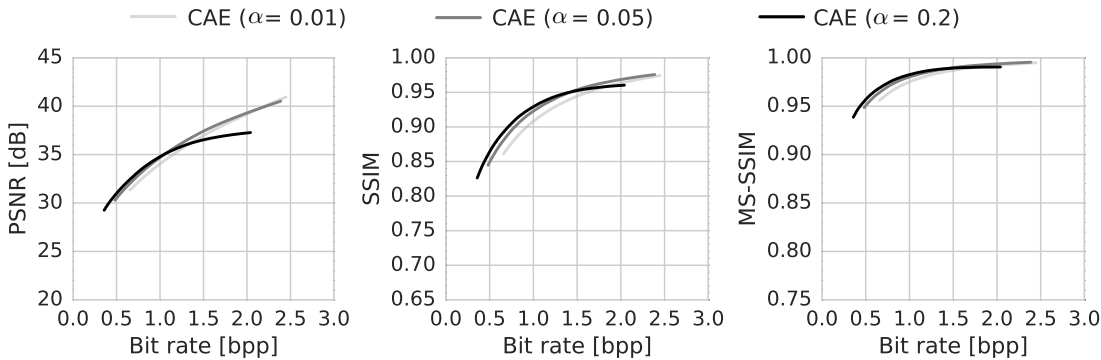

		To bring the parameter controlling the rate-distortion trade-off into a more intuitive
		range, we rescaled the distortion term and expressed the objective as follows:
		\begin{align}
			-\frac{\alpha}{N} \ln q\left( \left[f(\mathbf{x}) \circ \bm{\lambda} \right] + \mathbf{u} \right)
			+ \frac{1 - \alpha}{1000 \cdot M} \cdot ||\mathbf{x} - g( \left[f(\mathbf{x}) \circ \bm{\lambda} \right] / \bm{\lambda} ) ||^2.
		\end{align}
		Here, $N$ is the number of coefficients produced by the encoder and
		$M$ is the dimensionality of $\mathbf{x}$ (i.e., 3 times the number of pixels).

		The high-bit-rate CAE was trained with $\alpha = 0.01$ and 96 output channels, the medium-bit-rate CAE was trained with
		$\alpha = 0.05$ and 96 output channels, and the low-bit-rate CAE was trained with $\alpha = 0.2$ and 64 output channels.

	\subsection{Comparison with VAE}
		\label{sec:vae}
		\begin{figure}[H]
			\centering
			\hbox {
				\hspace{-1cm}
				\input{figures/quantitative_comparison_balle.tex}
			}
			\caption{An alternative to our approach is to replace the rounding function with additive uniform
			noise during training \citep{Balle:2016}. Using mean-squared error for measuring distortion, optimizing
			rate-distortion this way is equivalent to training a variational autoencoder \citep{Kingma:2014} with
			a Gaussian likelihood and uniform encoder (Section~\ref{sec:related}). Using the same training proecedure
			autoencoder architecture for both approaches (here trained for high bit-rates), we find
			that additive noise performs worse than redefining derivatives as in our approach.
			Rounding-based quantization is used at test time in both approaches.}
			\label{fig:vae}
		\end{figure}
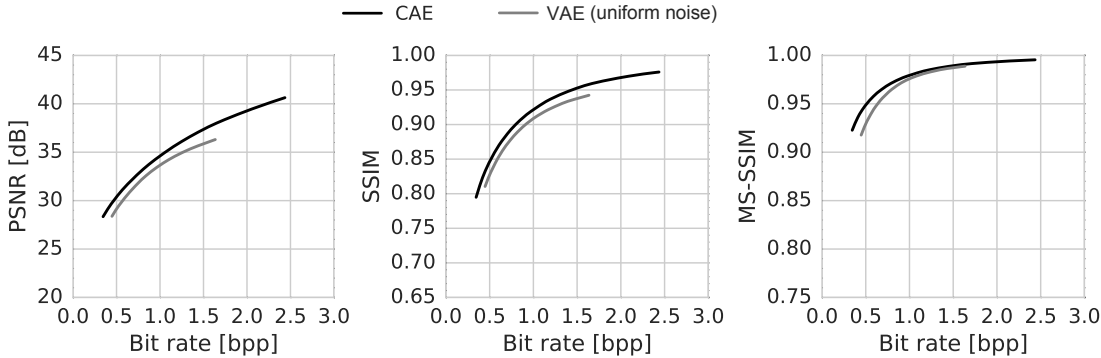

	\subsection{Complete images}
		\label{sec:details}
		Below we show complete images corresponding to the crops in Figure~\ref{fig:closeups}. For each
		image, we show the original image (top left), reconstructions using CAE (top right),
		reconstructions using JPEG 2000 (bottom left) and reconstructions using the method of
		\citep{Toderici:2016b} (bottom right). The images are best viewed on a monitor
		screen.

		\begin{figure}[p]
			\hbox{
				\hspace{-2.5cm}
				\begin{tikzpicture}
					\node at (0.0cm, 6.2cm) {\includegraphics[width=9cm]{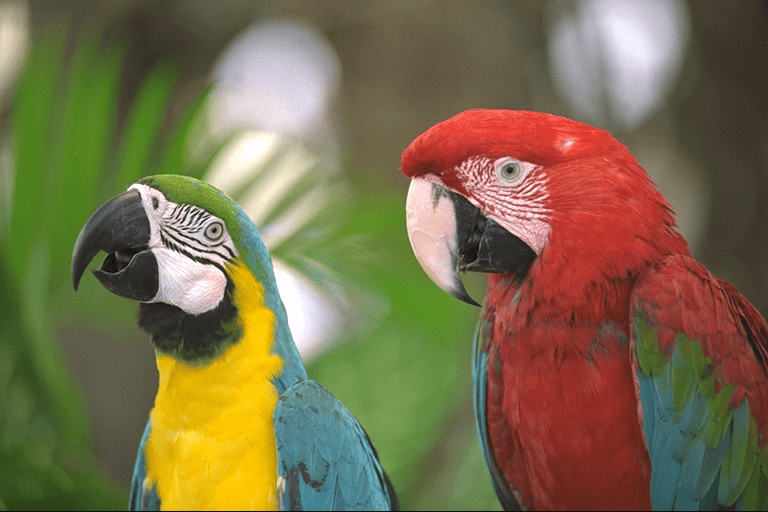}};
					\node at (9.2cm, 6.2cm) {\includegraphics[width=9cm]{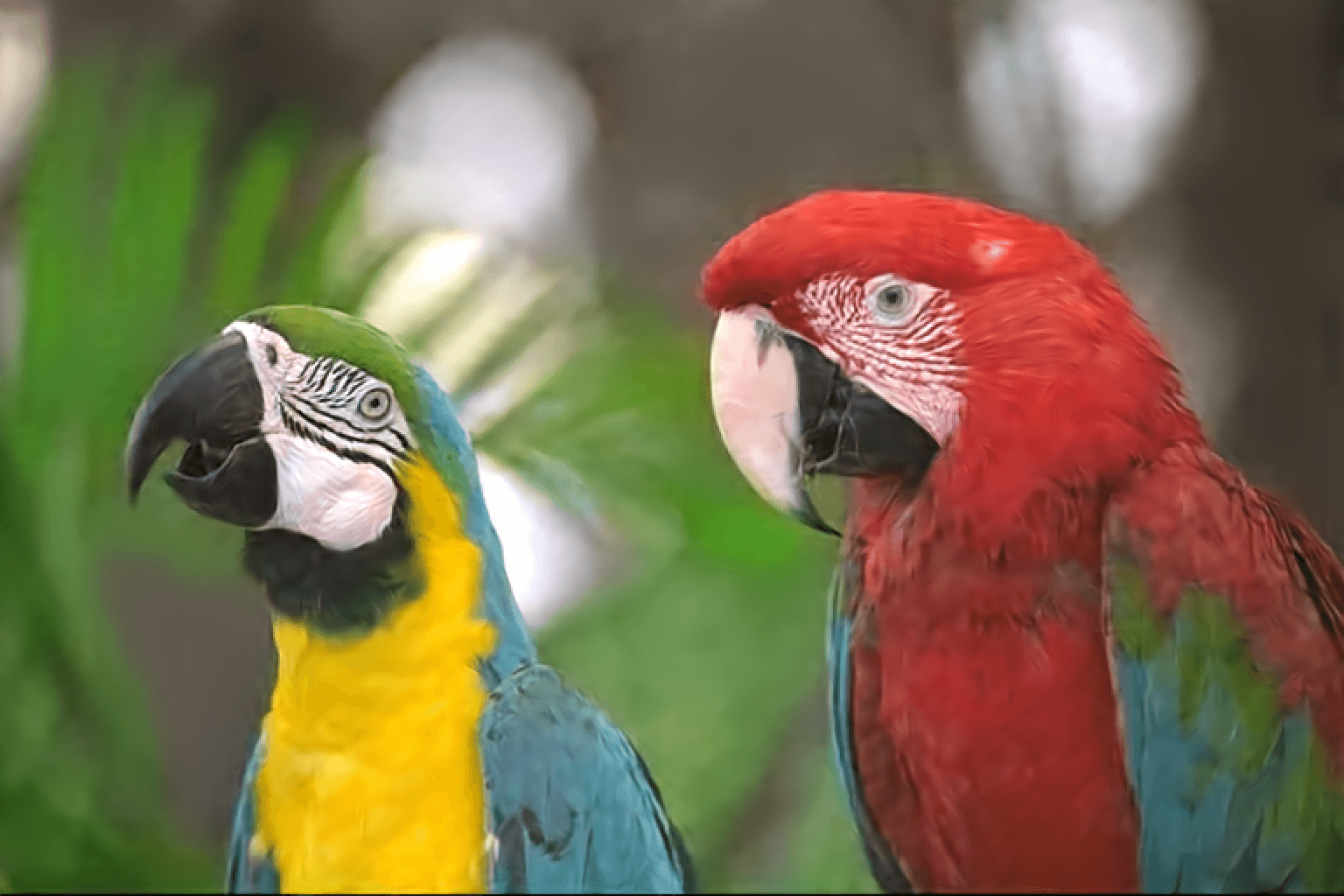}};
					\node at (0.0cm, 0.0cm) {\includegraphics[width=9cm]{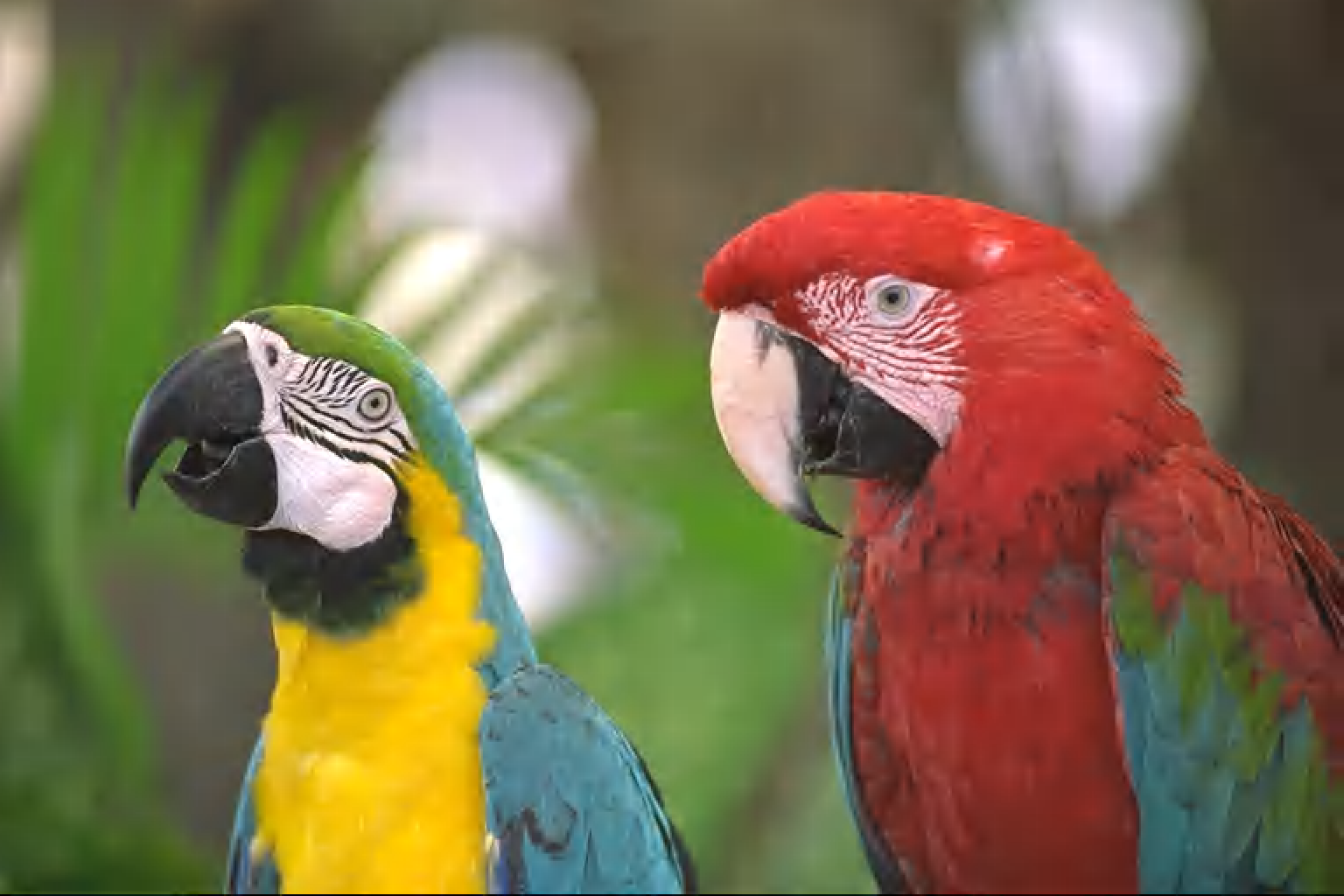}};
					\node at (9.2cm, 0.0cm) {\includegraphics[width=9cm]{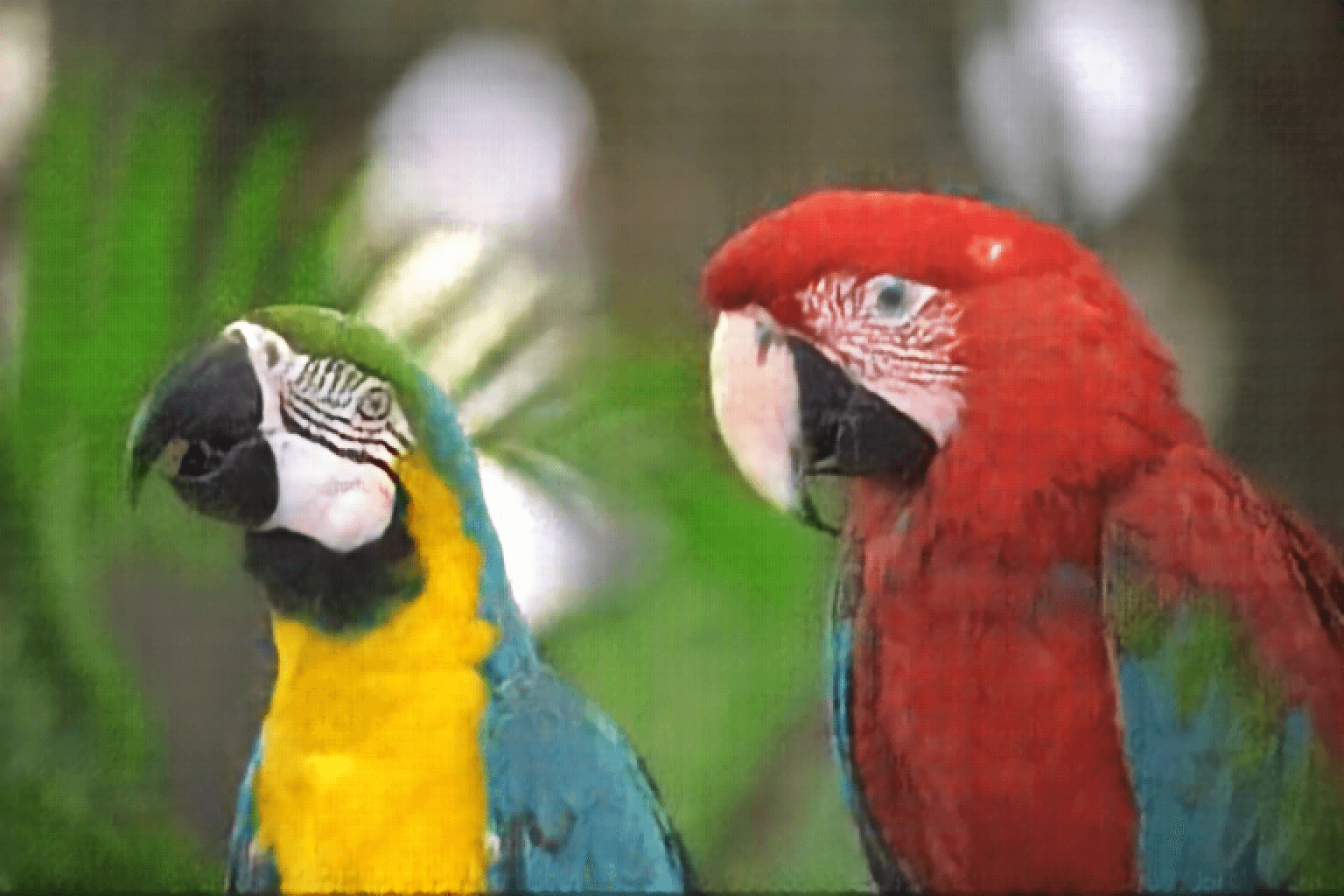}};
				\end{tikzpicture}
			}
		\end{figure}

		\begin{figure}[p]
			\vspace{-1cm}
			\hbox{
				\hspace{-1cm}
				\begin{tikzpicture}
					\node at (0.0cm, 12.2cm) {\includegraphics[width=8cm]{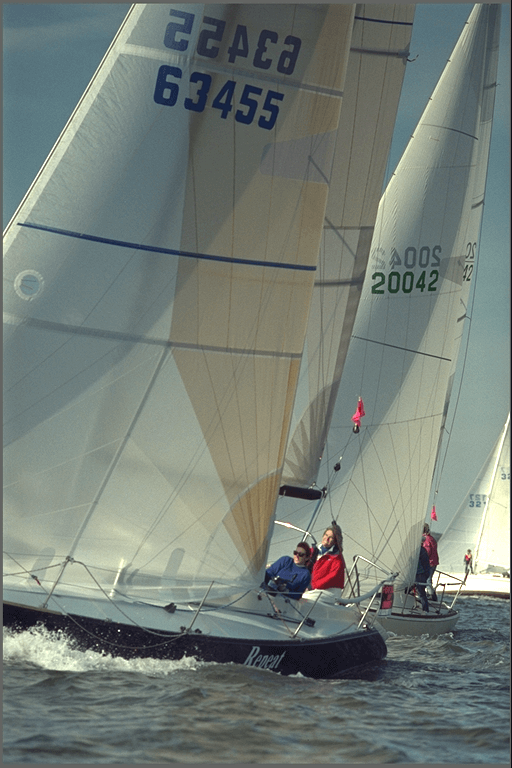}};
					\node at (8.2cm, 12.2cm) {\includegraphics[width=8cm]{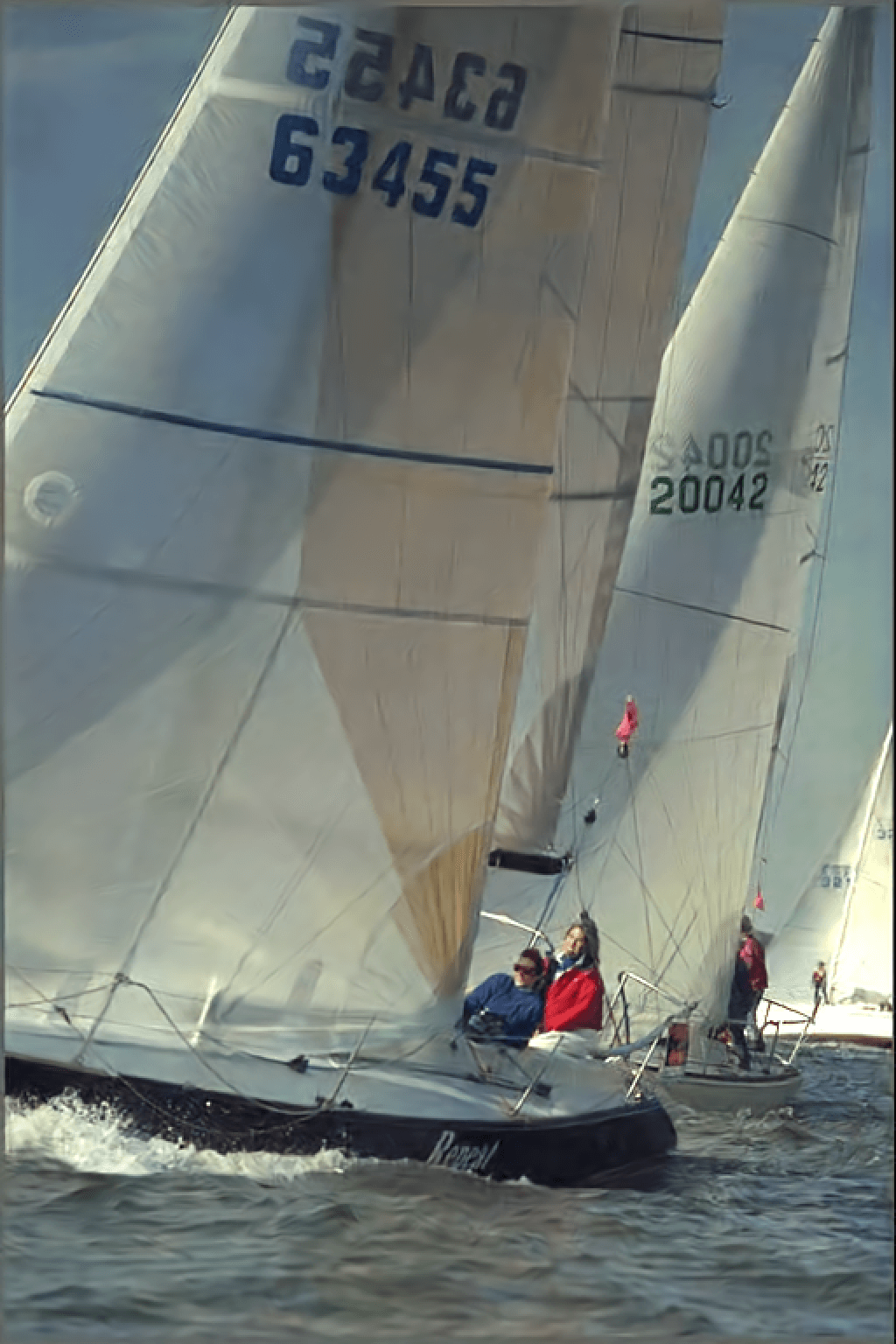}};
					\node at (0.0cm, 0.0cm) {\includegraphics[width=8cm]{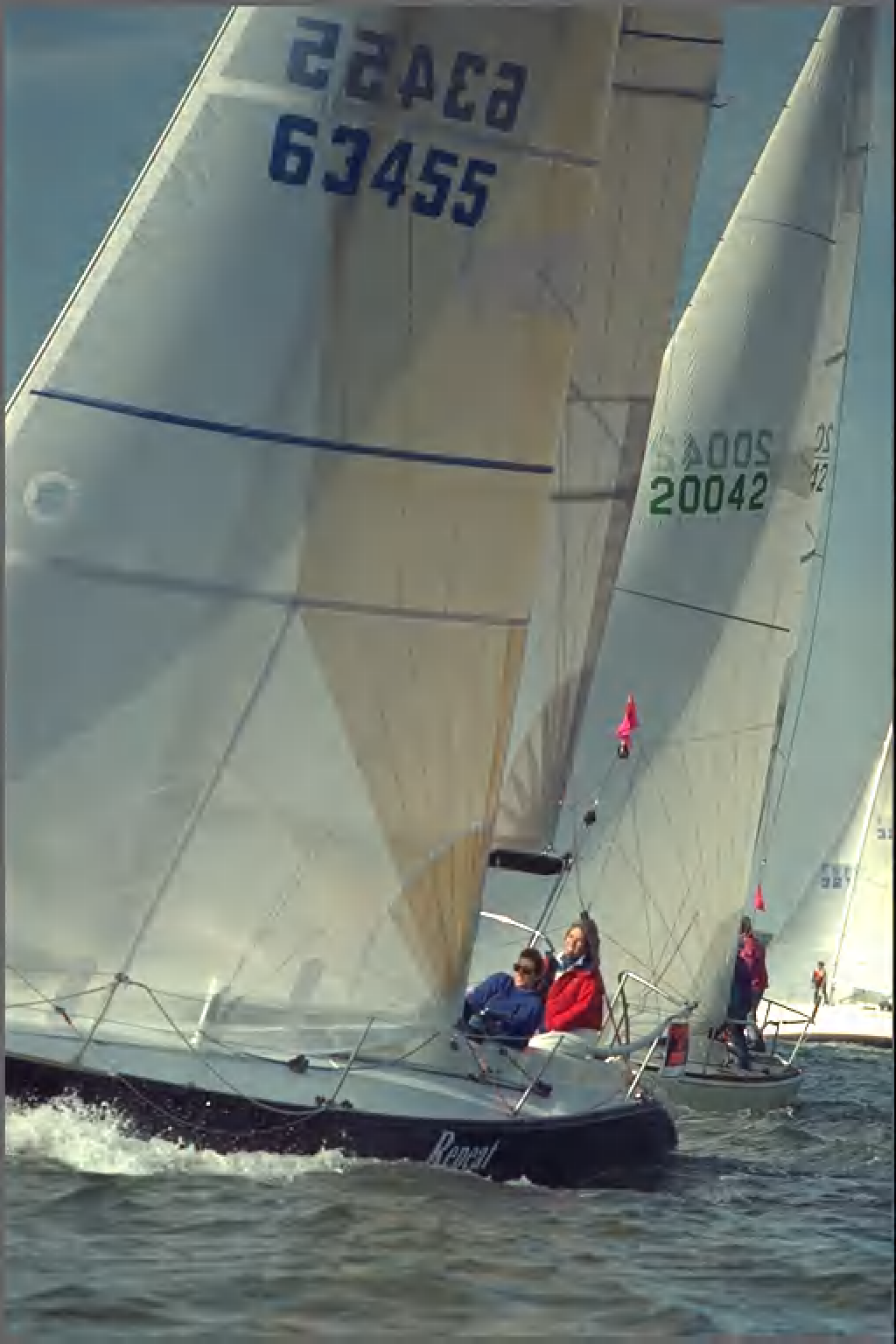}};
					\node at (8.2cm, 0.0cm) {\includegraphics[width=8cm]{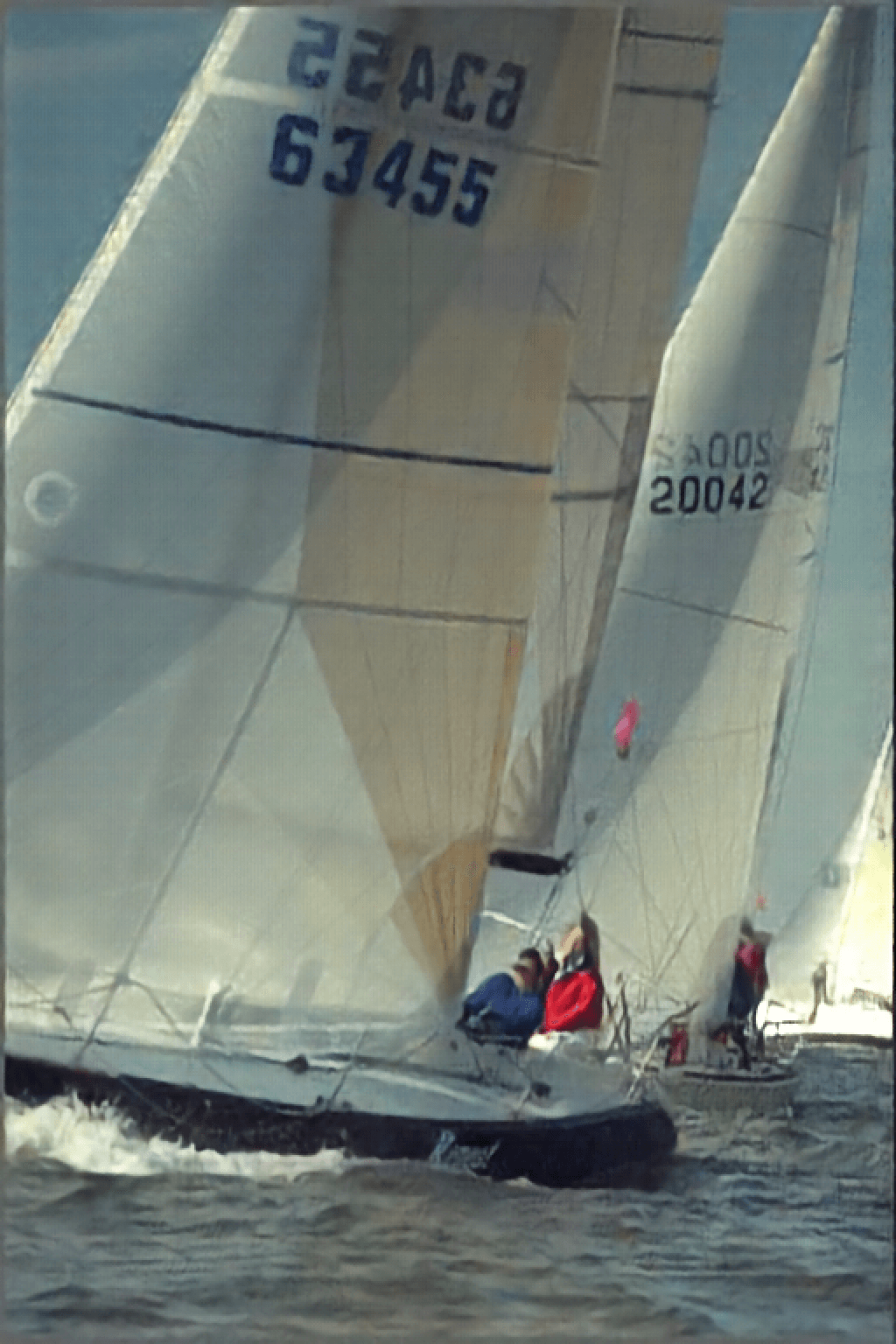}};
				\end{tikzpicture}
			}
		\end{figure}

		\begin{figure}[p]
			\vspace{-1cm}
			\hbox{
				\hspace{-1cm}
				\begin{tikzpicture}
					\node at (0.0cm, 12.2cm) {\includegraphics[width=8cm]{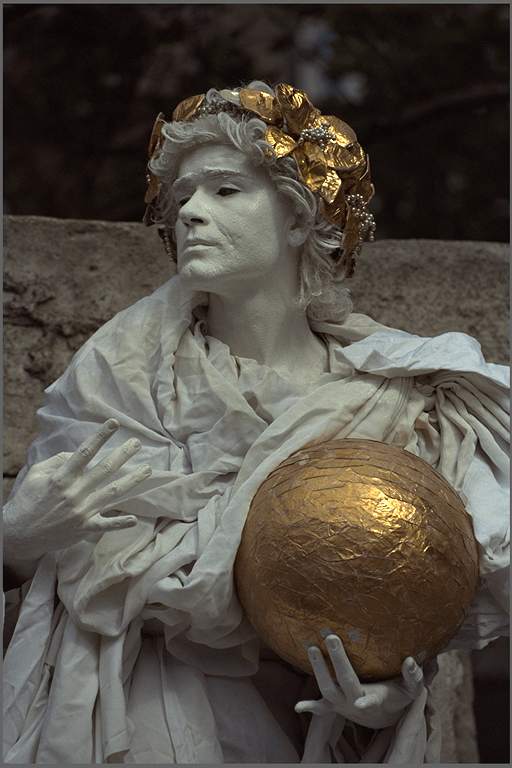}};
					\node at (8.2cm, 12.2cm) {\includegraphics[width=8cm]{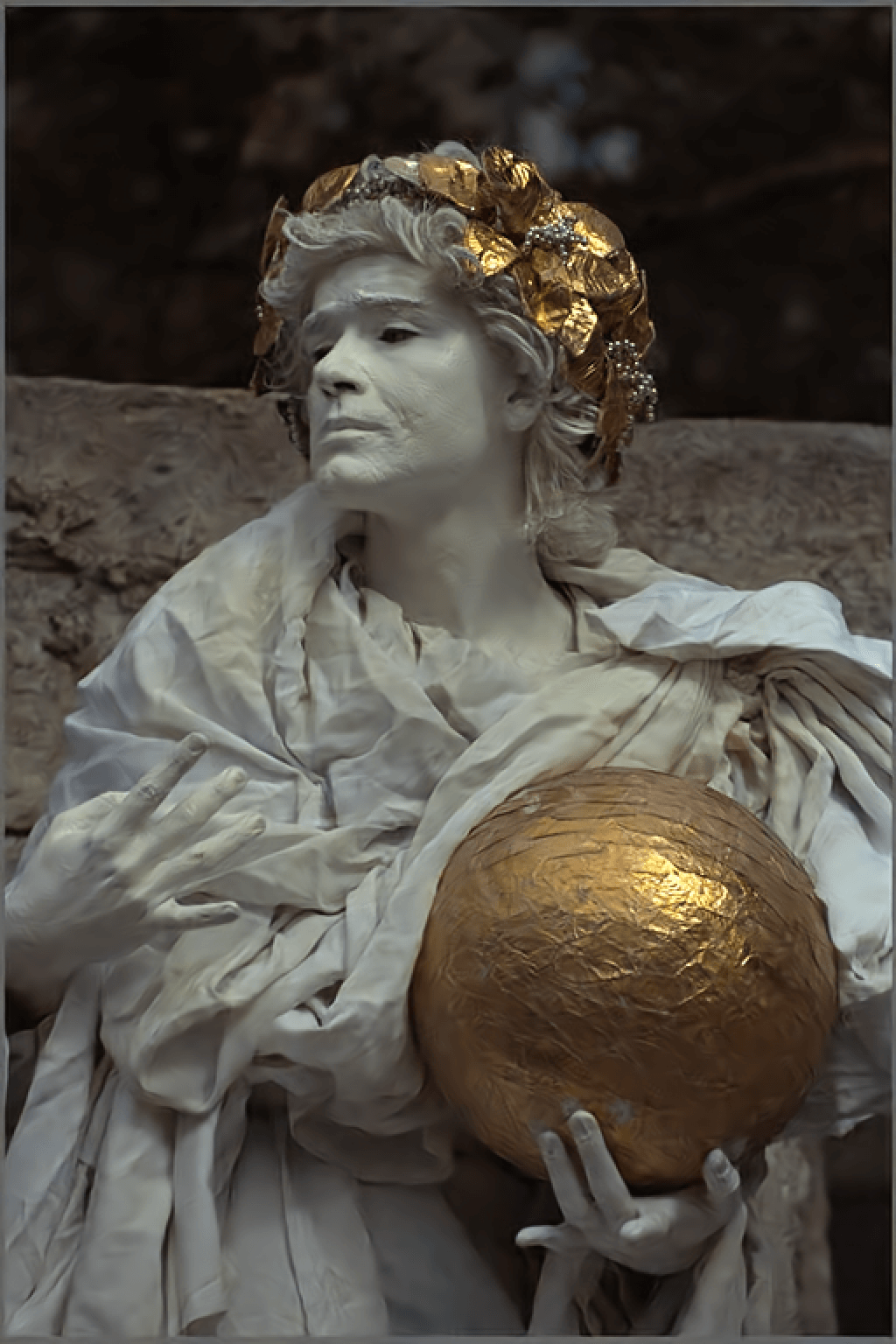}};
					\node at (0.0cm, 0.0cm) {\includegraphics[width=8cm]{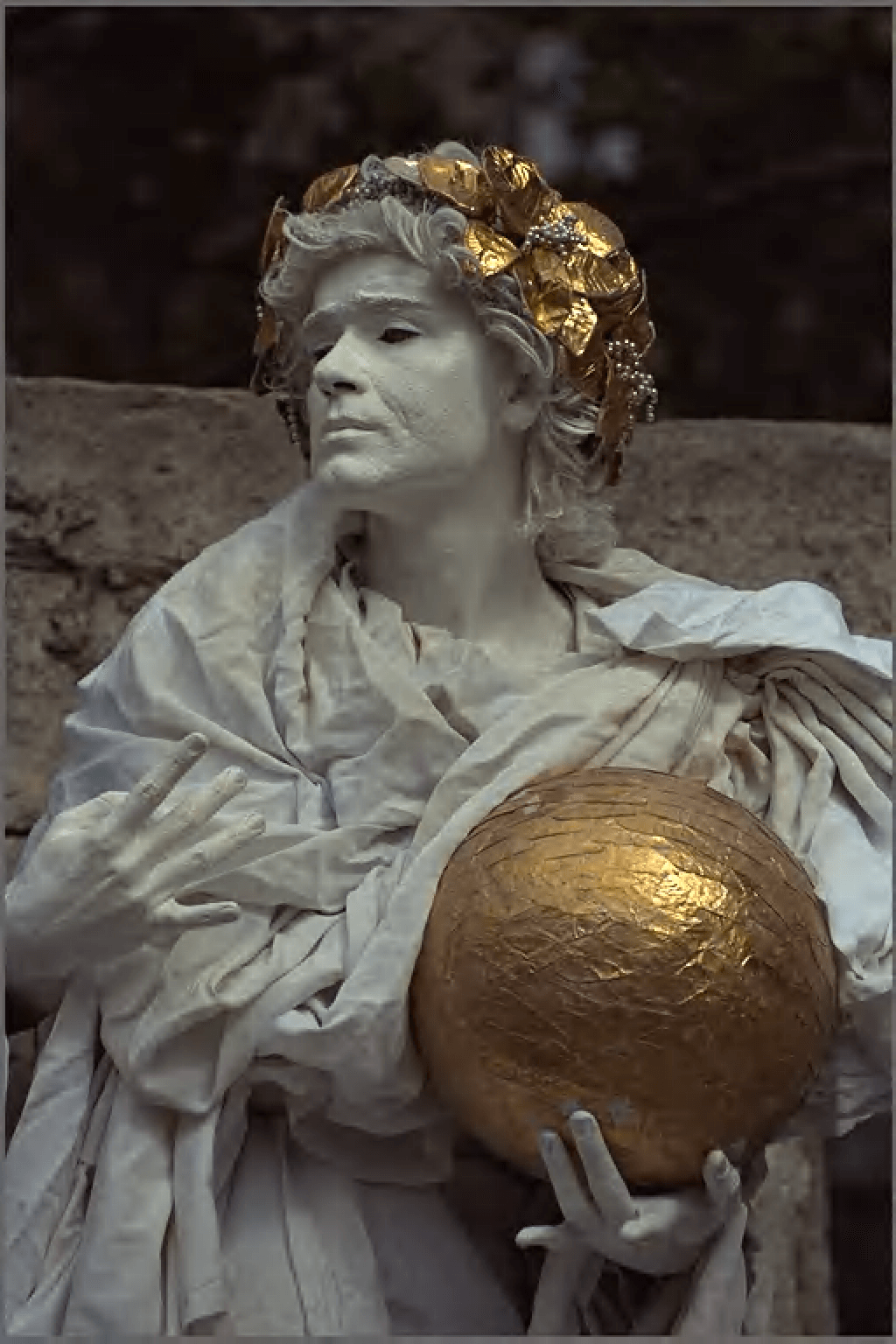}};
					\node at (8.2cm, 0.0cm) {\includegraphics[width=8cm]{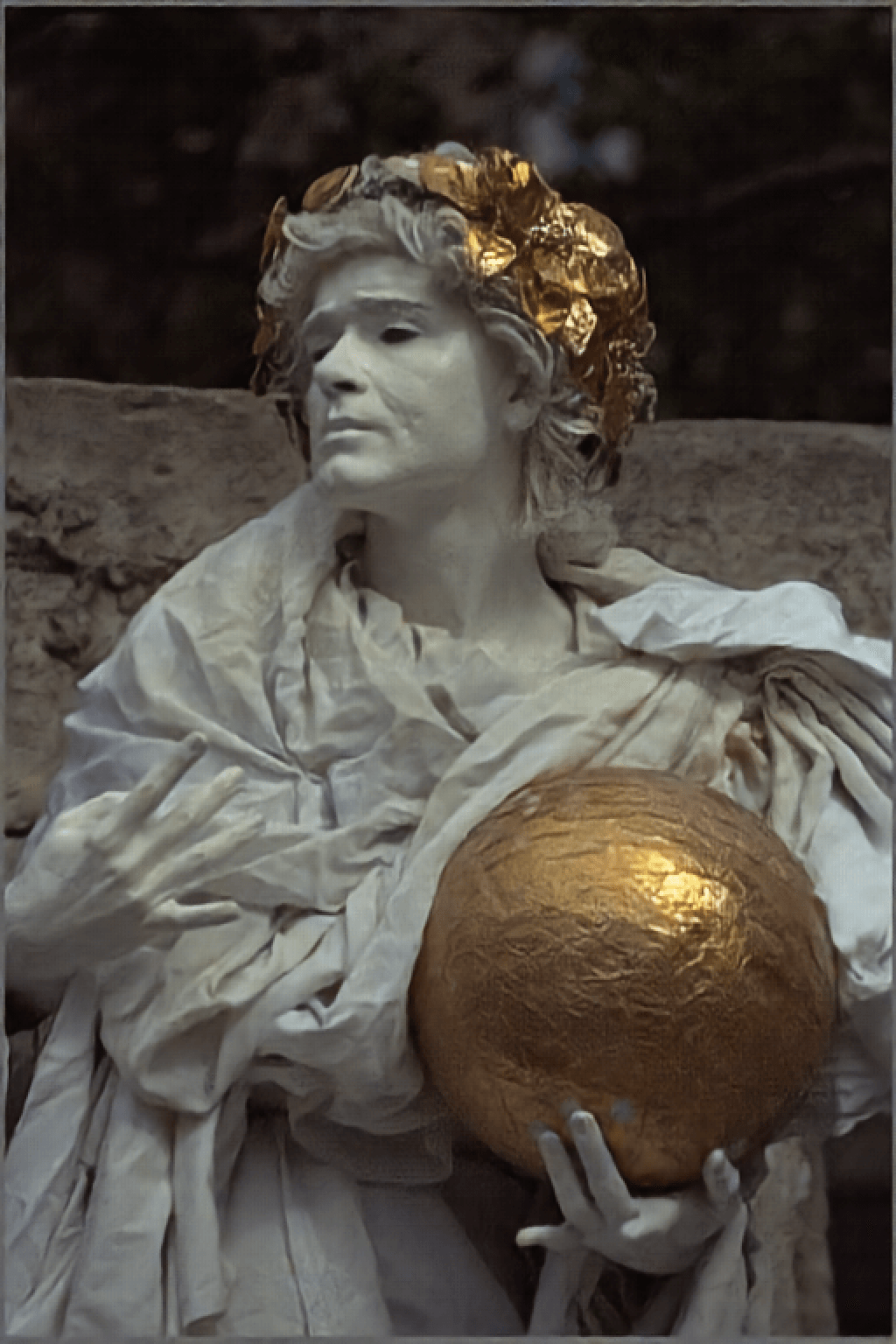}};
				\end{tikzpicture}
			}
		\end{figure}

		\begin{figure}[p]
			\vspace{-1cm}
			\hbox{
				\hspace{-1cm}
				\begin{tikzpicture}
					\node at (0.0cm, 12.2cm) {\includegraphics[width=8cm]{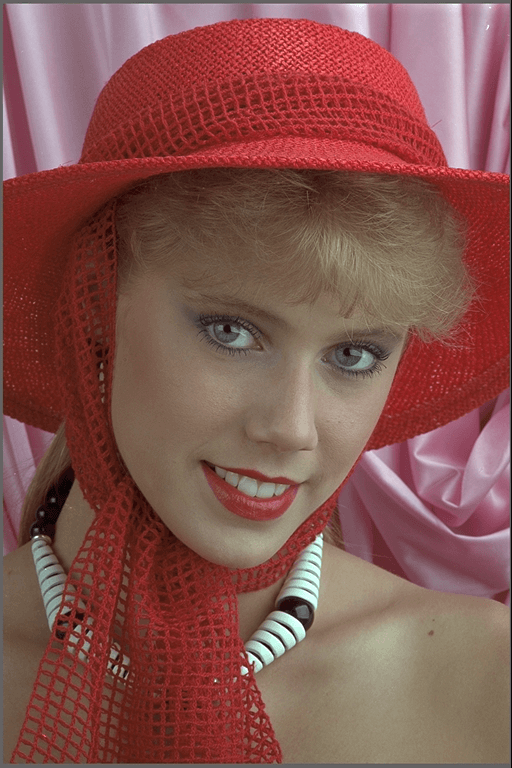}};
					\node at (8.2cm, 12.2cm) {\includegraphics[width=8cm]{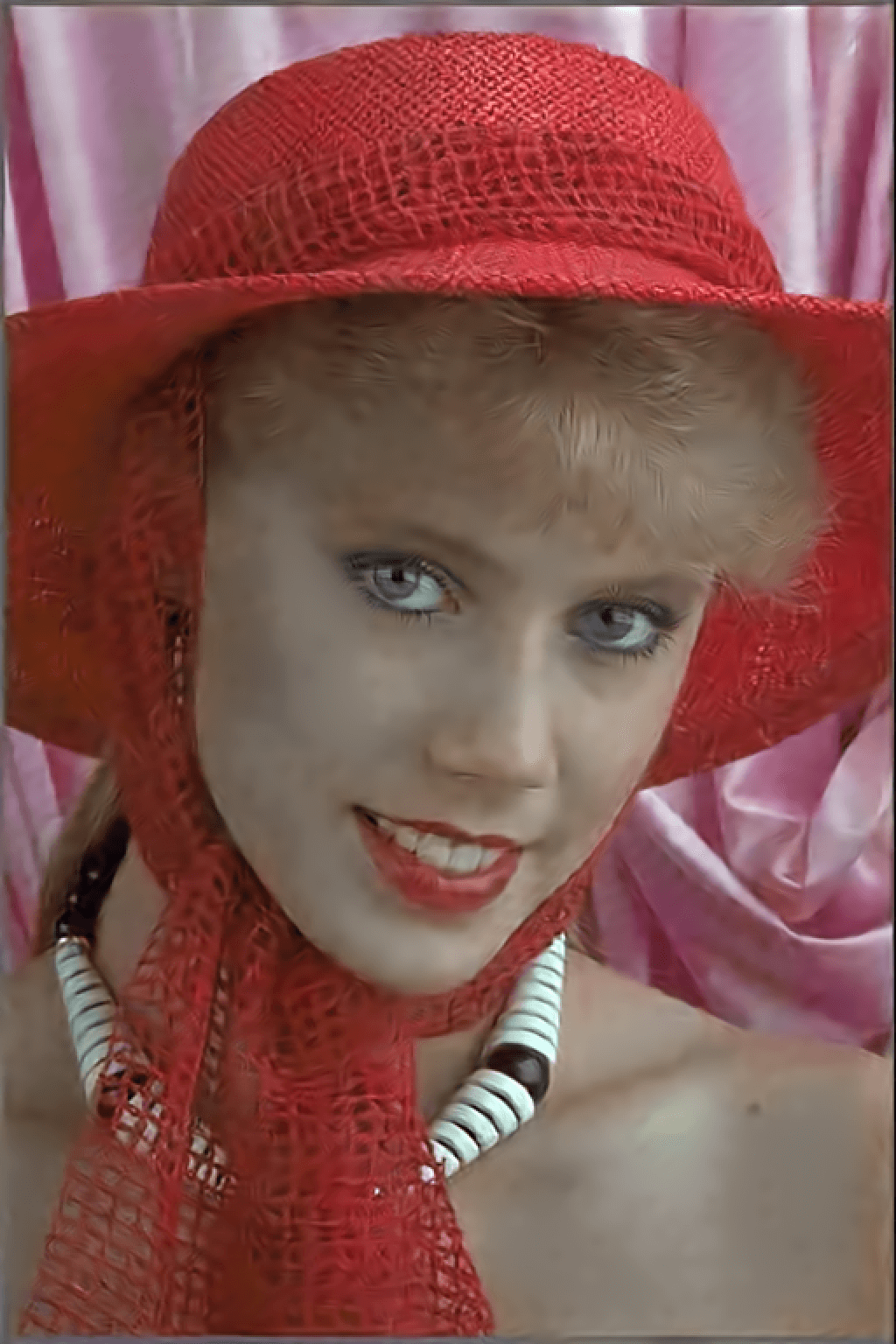}};
					\node at (0.0cm, 0.0cm) {\includegraphics[width=8cm]{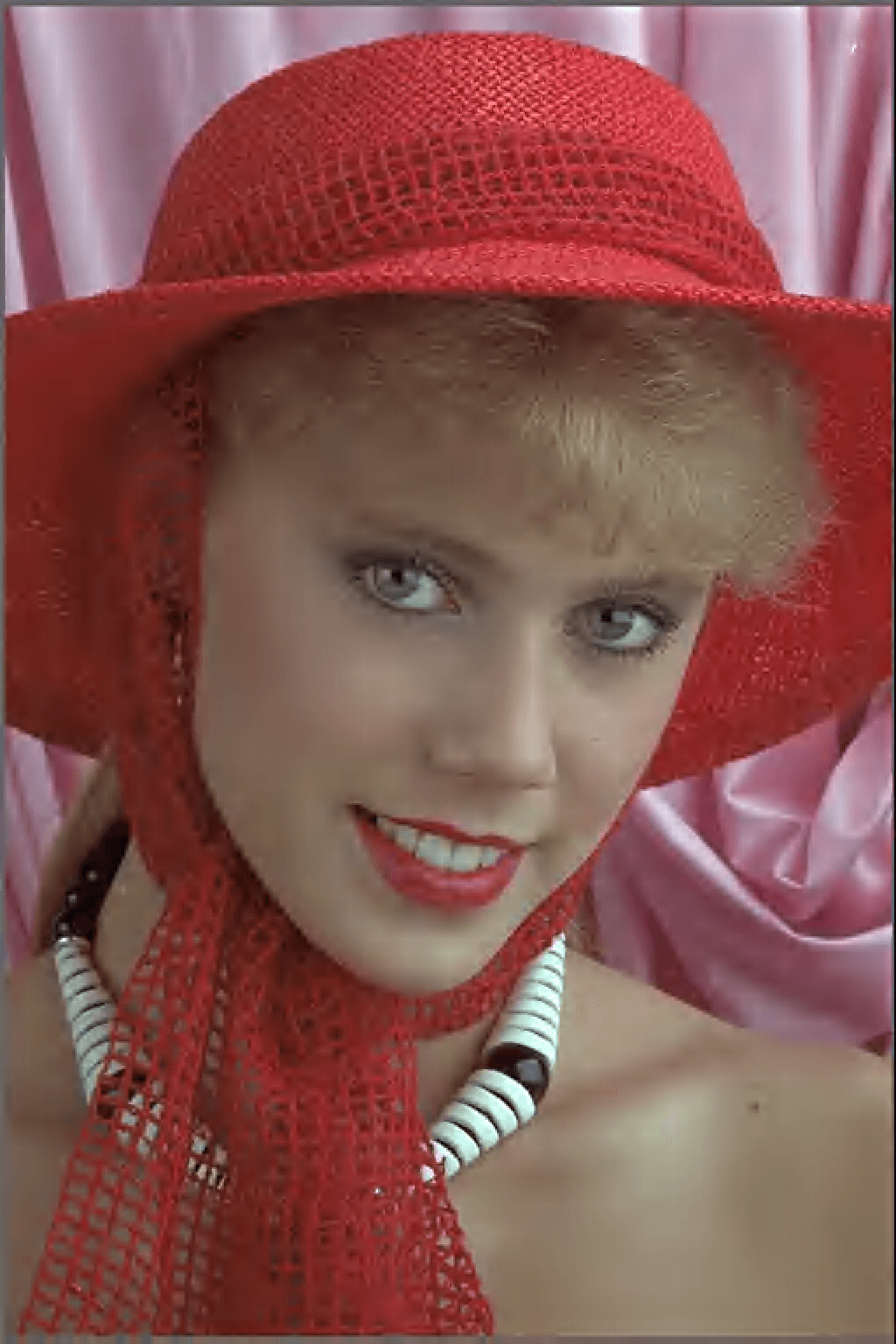}};
					\node at (8.2cm, 0.0cm) {\includegraphics[width=8cm]{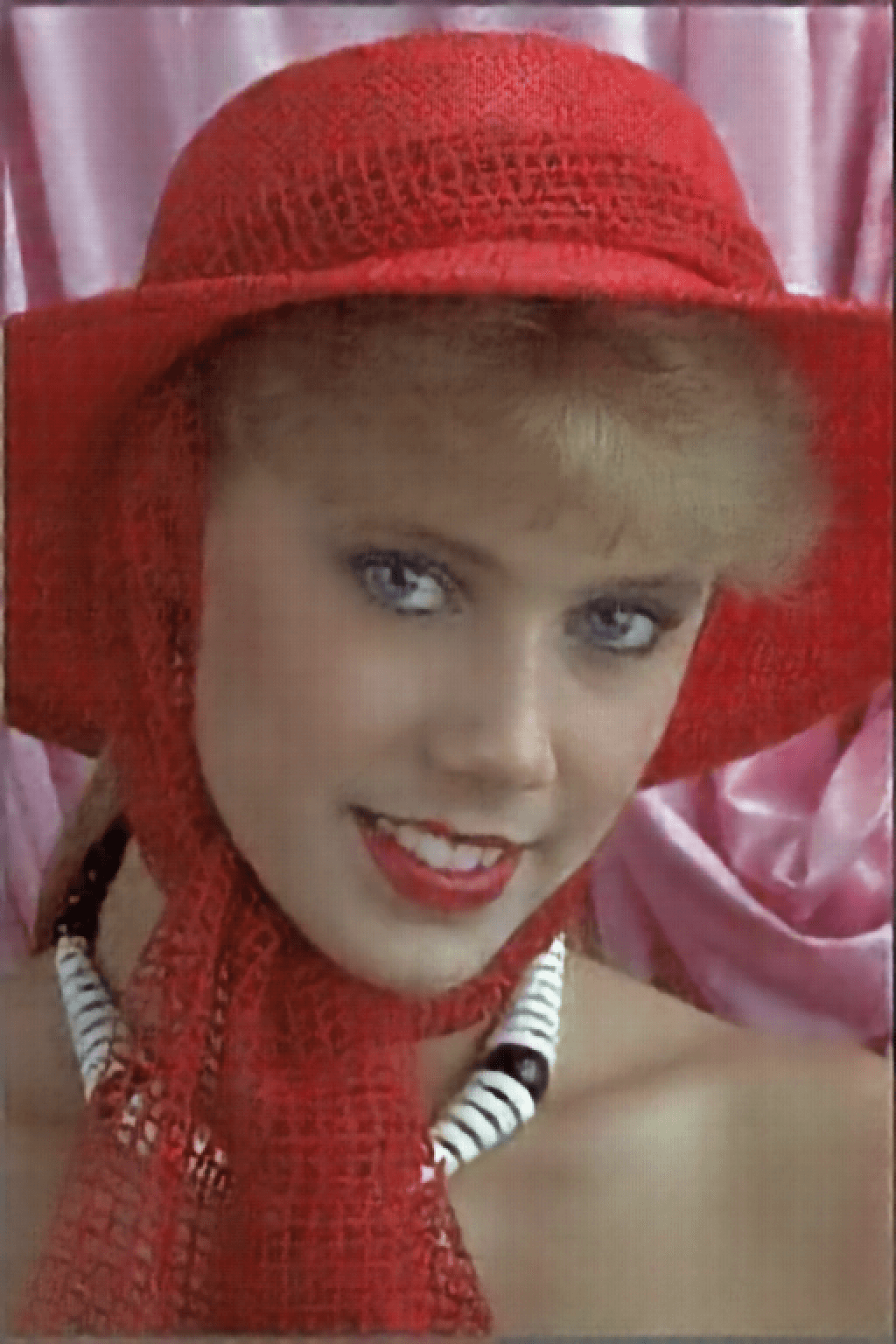}};
				\end{tikzpicture}
			}
		\end{figure}

		\begin{figure}[p]
			\vspace{-1cm}
			\hbox{
				\hspace{-1cm}
				\begin{tikzpicture}
					\node at (0.0cm, 12.2cm) {\includegraphics[width=8cm]{figures/images/kodim04.png}};
					\node at (8.2cm, 12.2cm) {\includegraphics[width=8cm]{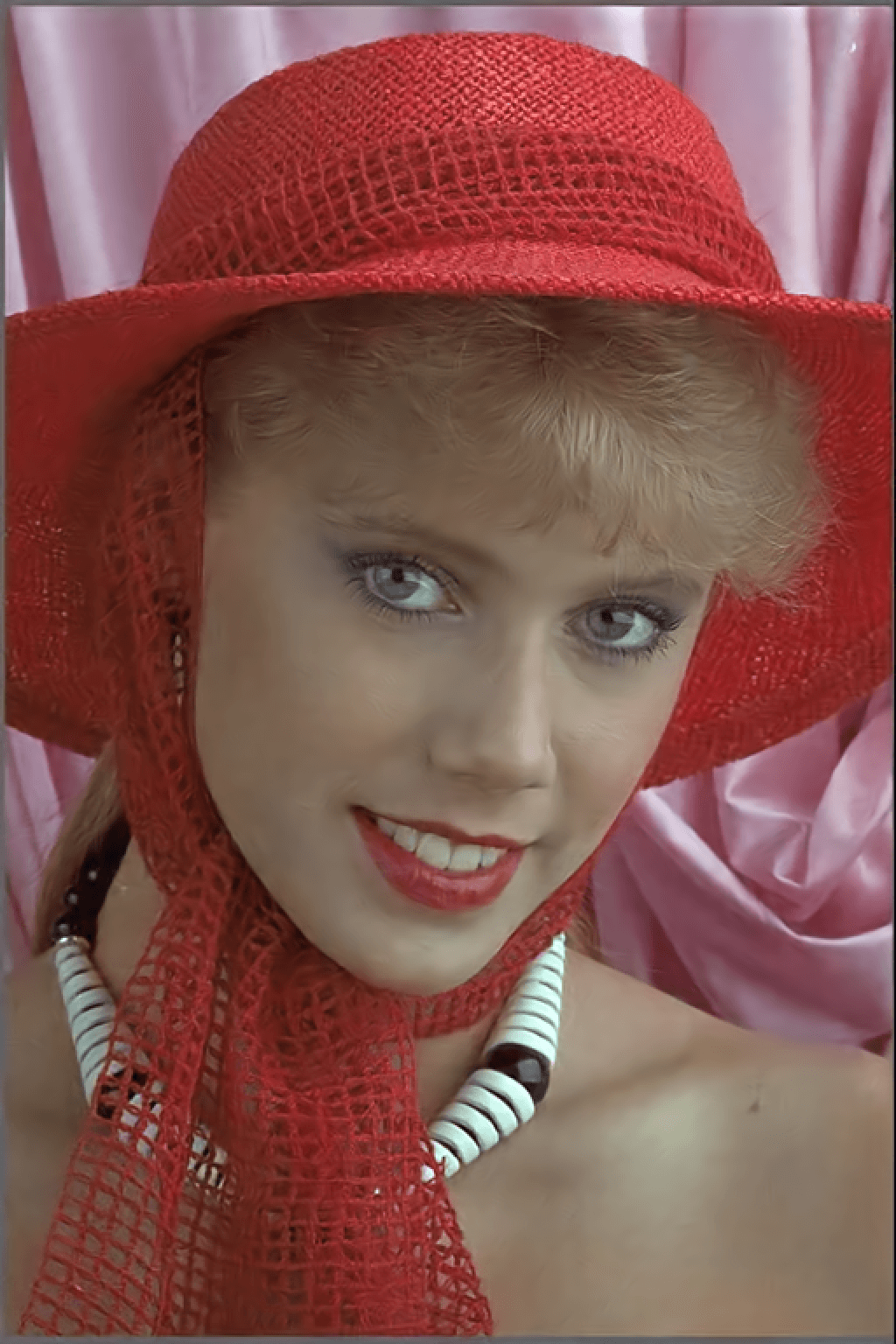}};
					\node at (0.0cm, 0.0cm) {\includegraphics[width=8cm]{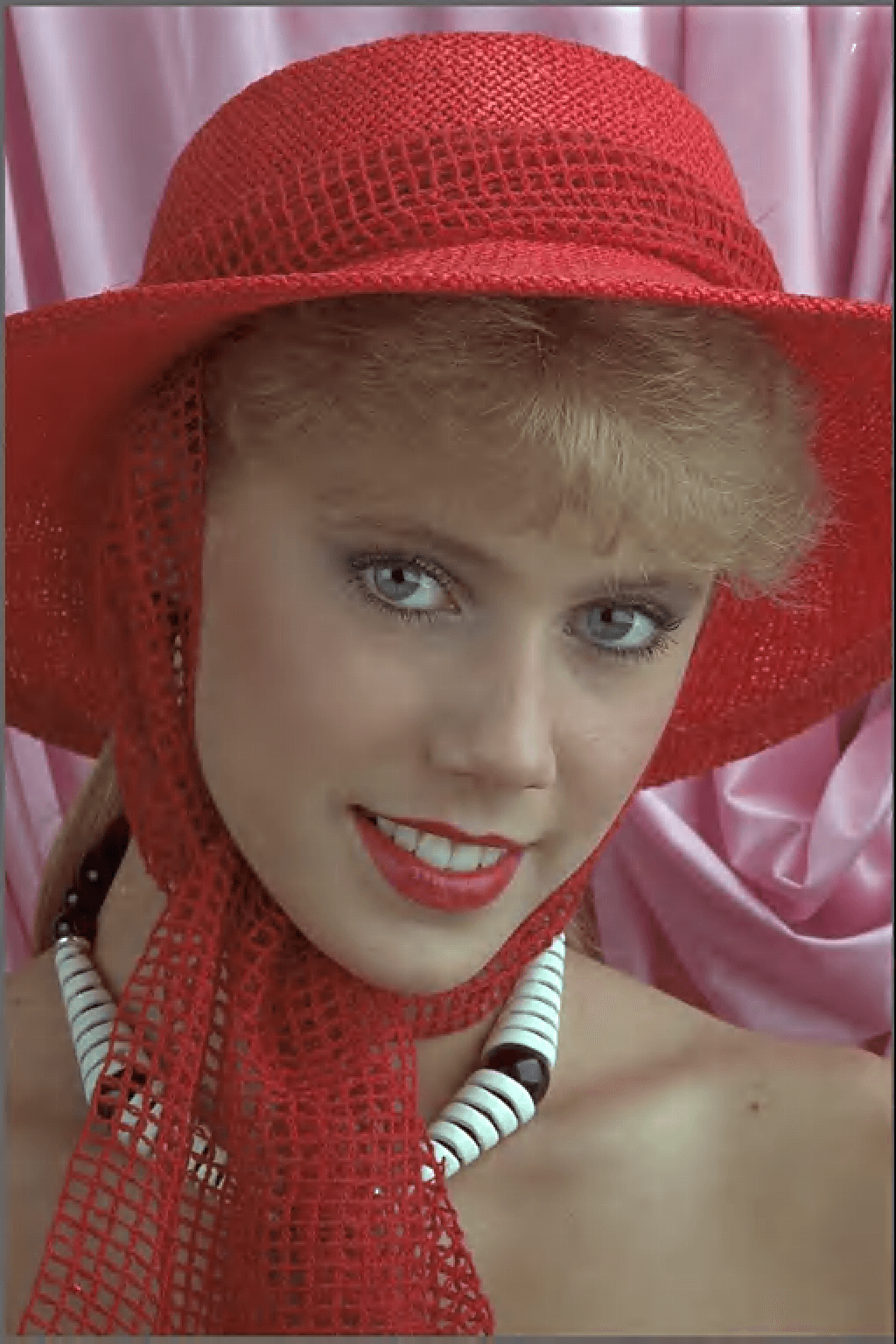}};
					\node at (8.2cm, 0.0cm) {\includegraphics[width=8cm]{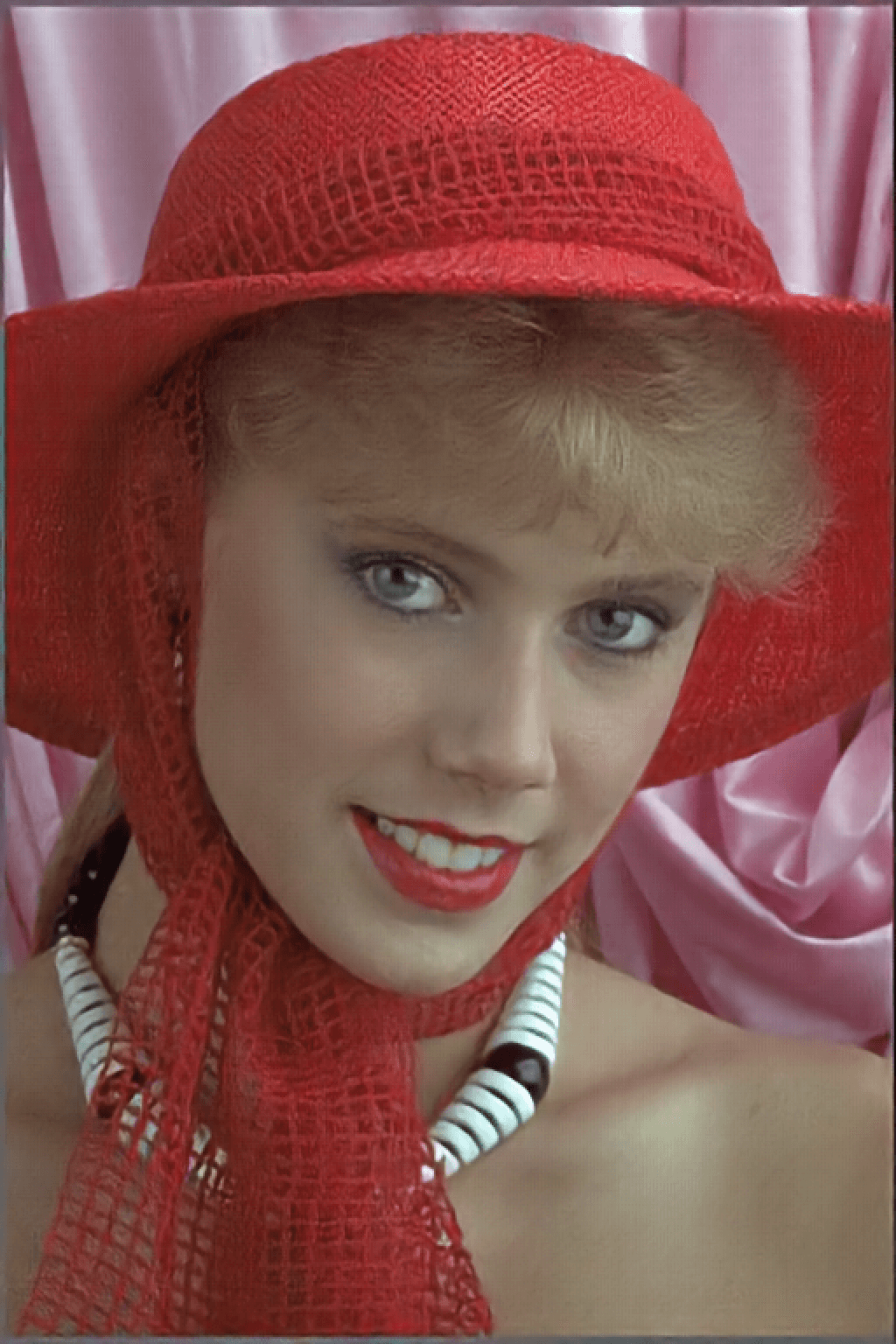}};
				\end{tikzpicture}
			}
		\end{figure}
\end{document}

%% file: figures/quantization.tex
\begin{tikzpicture}
	\node at (-6.9cm, 2cm) {\footnotesize\textsf{\textbf{A}}};
	\node at (-3.4cm, 2cm) {\footnotesize\textsf{\textbf{B}}};
	\node at ( 0.1cm, 2cm) {\footnotesize\textsf{\textbf{C}}};
	\node at ( 3.6cm, 2cm) {\footnotesize\textsf{\textbf{D}}};

	\node at (-5.2cm, 2cm) {\scriptsize\textsf{Original}};
	\node at (-1.7cm, 2cm) {\scriptsize\textsf{Rounding}};
	\node at ( 1.8cm, 2cm) {\scriptsize\textsf{Stochastic rounding}};
	\node at ( 5.3cm, 2cm) {\scriptsize\textsf{Additive noise}};

	\node at (0, 0) {\includegraphics[width=14cm]{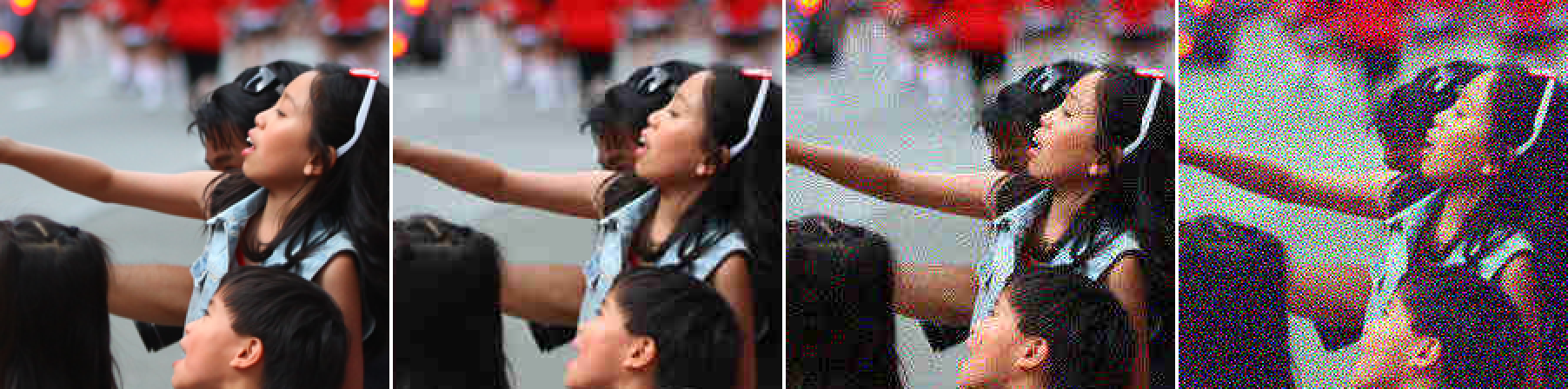}};
\end{tikzpicture}

%% file: figures/training.tex
\begin{tikzpicture}
	\node at (0cm, 0cm) {\includegraphics[height=3.5cm]{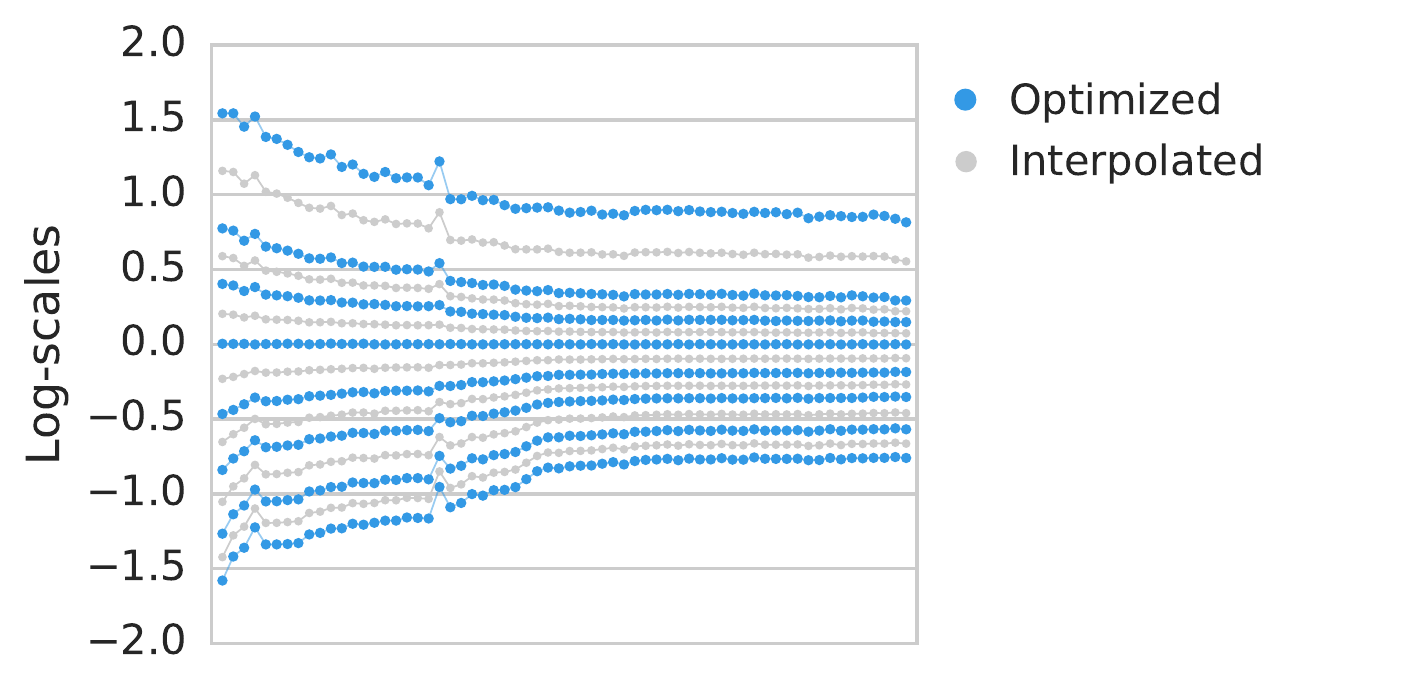}};
	\node at (7.3cm, -.22cm) {\includegraphics[height=3.9cm]{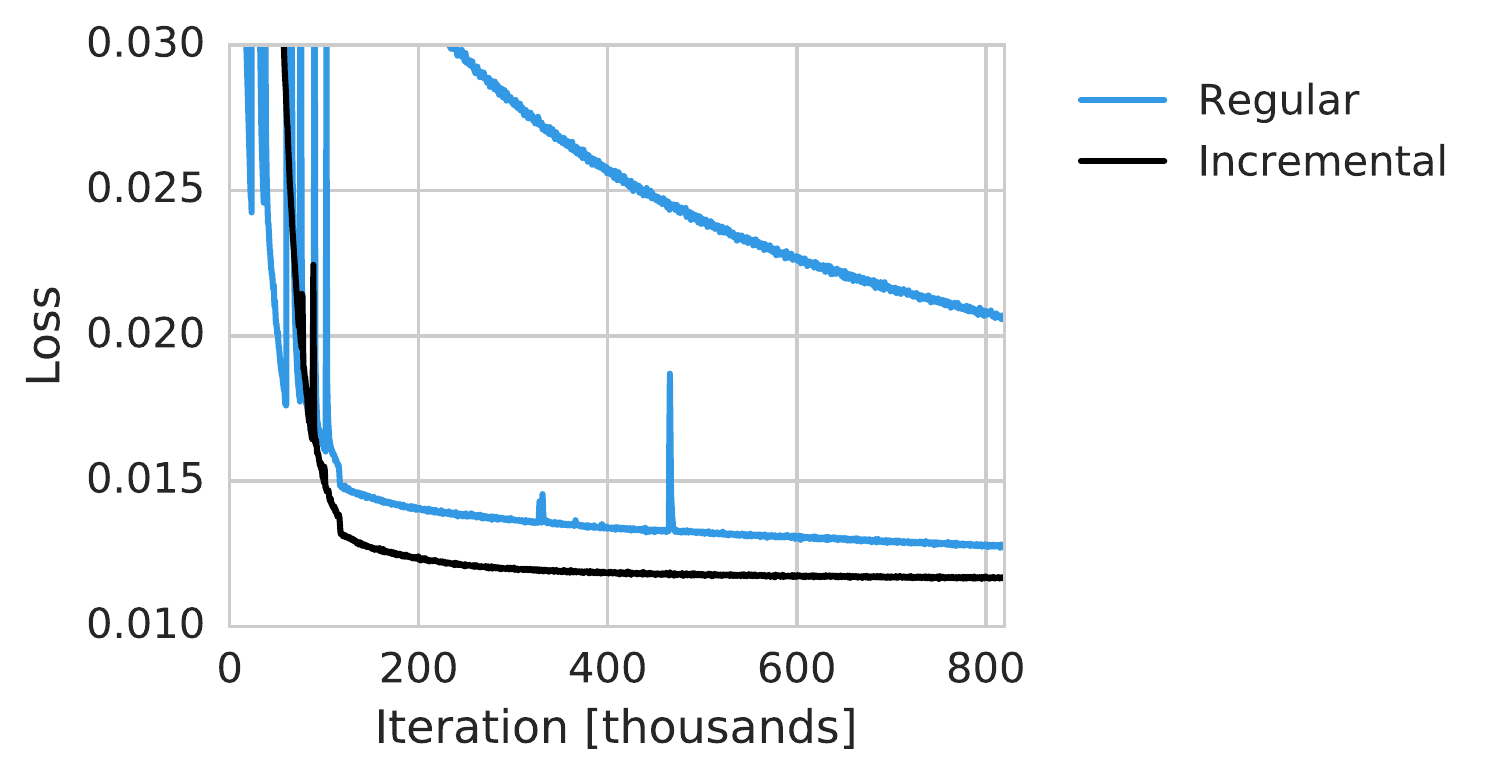}};

	\node at (-3.4cm, 2.1cm) {\footnotesize\textsf{\textbf{A}}};
	\node at ( 3.6cm, 2.1cm) {\footnotesize\textsf{\textbf{B}}};
\end{tikzpicture}

%% file: figures/quantitative_comparison.tex
\begin{tikzpicture}
	\node at (3.0cm, 2.5cm) {\includegraphics[width=13cm]{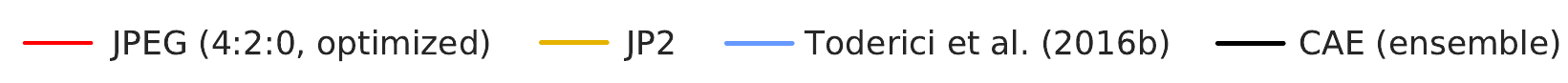}};

	\node[anchor=west] at (-4.7cm, 0cm) {\includegraphics[height=4.4cm]{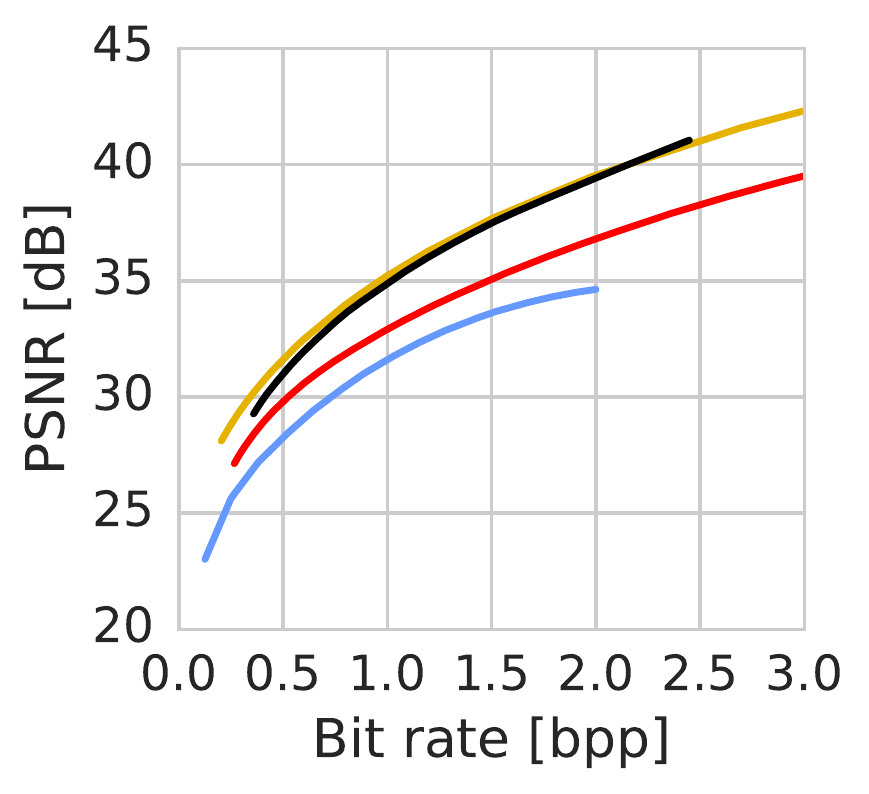}};
	\node[anchor=west] at ( 0.0cm, 0cm) {\includegraphics[height=4.4cm]{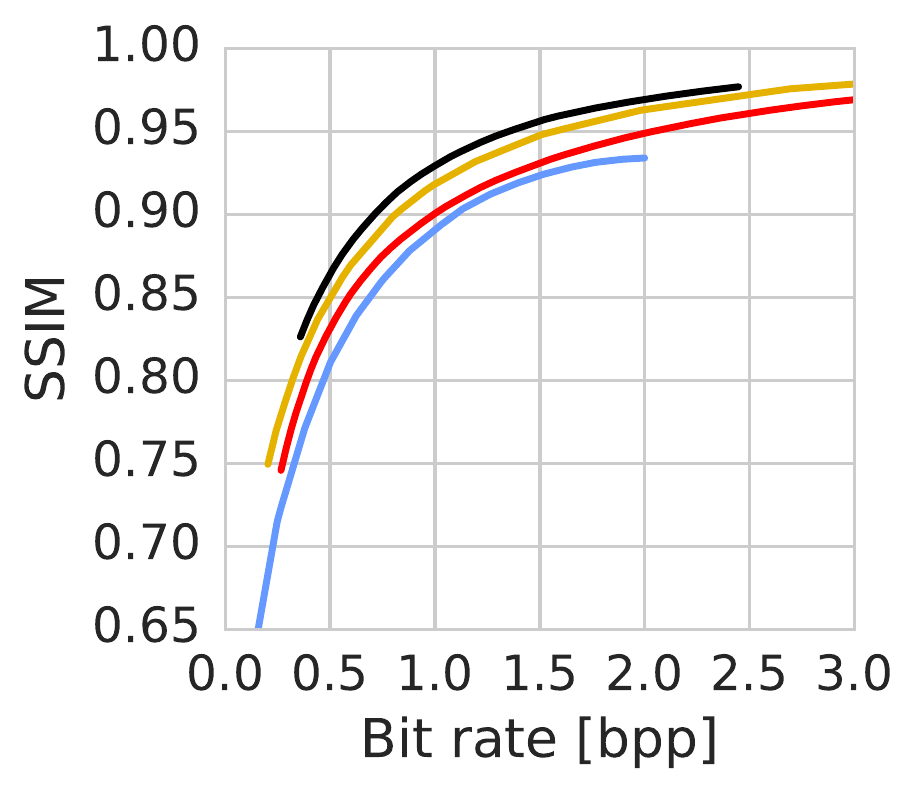}};
	\node[anchor=west] at ( 5.0cm, 0cm) {\includegraphics[height=4.4cm]{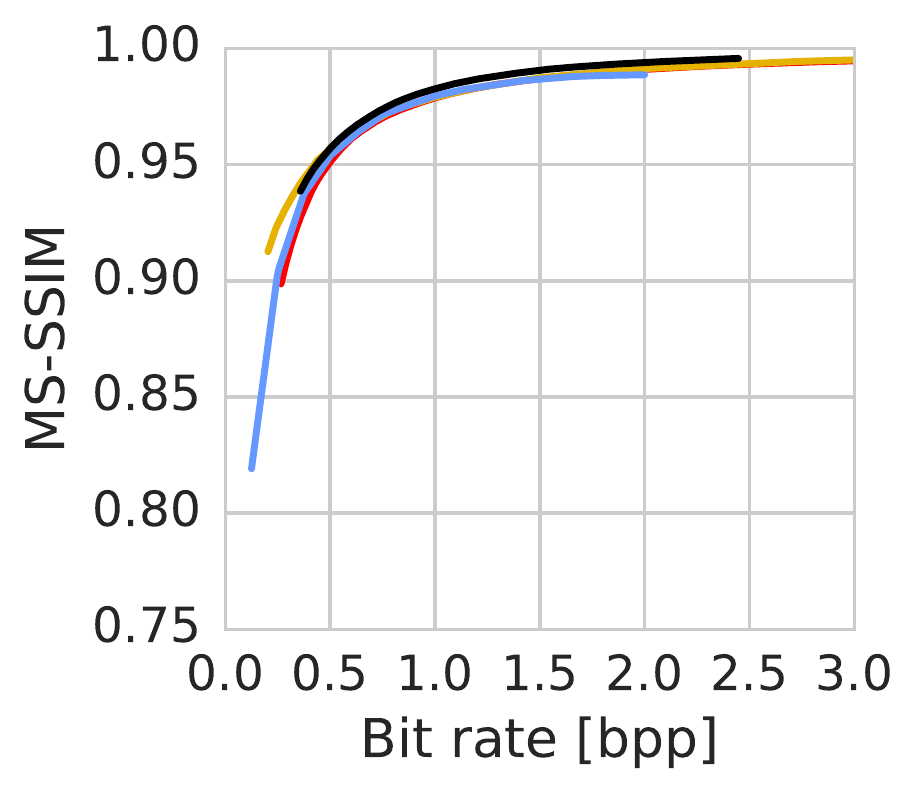}};
\end{tikzpicture}

%% file: figures/closeups.tex
\begin{tikzpicture}
	\node at ( 0cm, 2cm) {\scriptsize\textsf{CAE}};
	\node at ( 4cm, 2cm) {\scriptsize\textsf{JPEG 2000}};
	\node at ( 8cm, 2cm) {\scriptsize\textsf{JPEG}};
	\node at (12cm, 2cm) {\scriptsize\textsf{Toderici et al. (2016b)}};

	\node at ( 0cm, 0cm) {\includegraphics[trim={1cm 2.5cm 8cm 2.5cm},clip,resolution=302]{figures/images/kodim23_0.25bpp_cae.png}};
	\node at ( 4cm, 0cm) {\includegraphics[trim={1cm 2.5cm 8cm 2.5cm},clip,resolution=302]{figures/images/kodim23_0.25bpp_jp2.png}};
	\node at ( 8cm, 0cm) {\includegraphics[trim={1cm 2.5cm 8cm 2.5cm},clip,resolution=302]{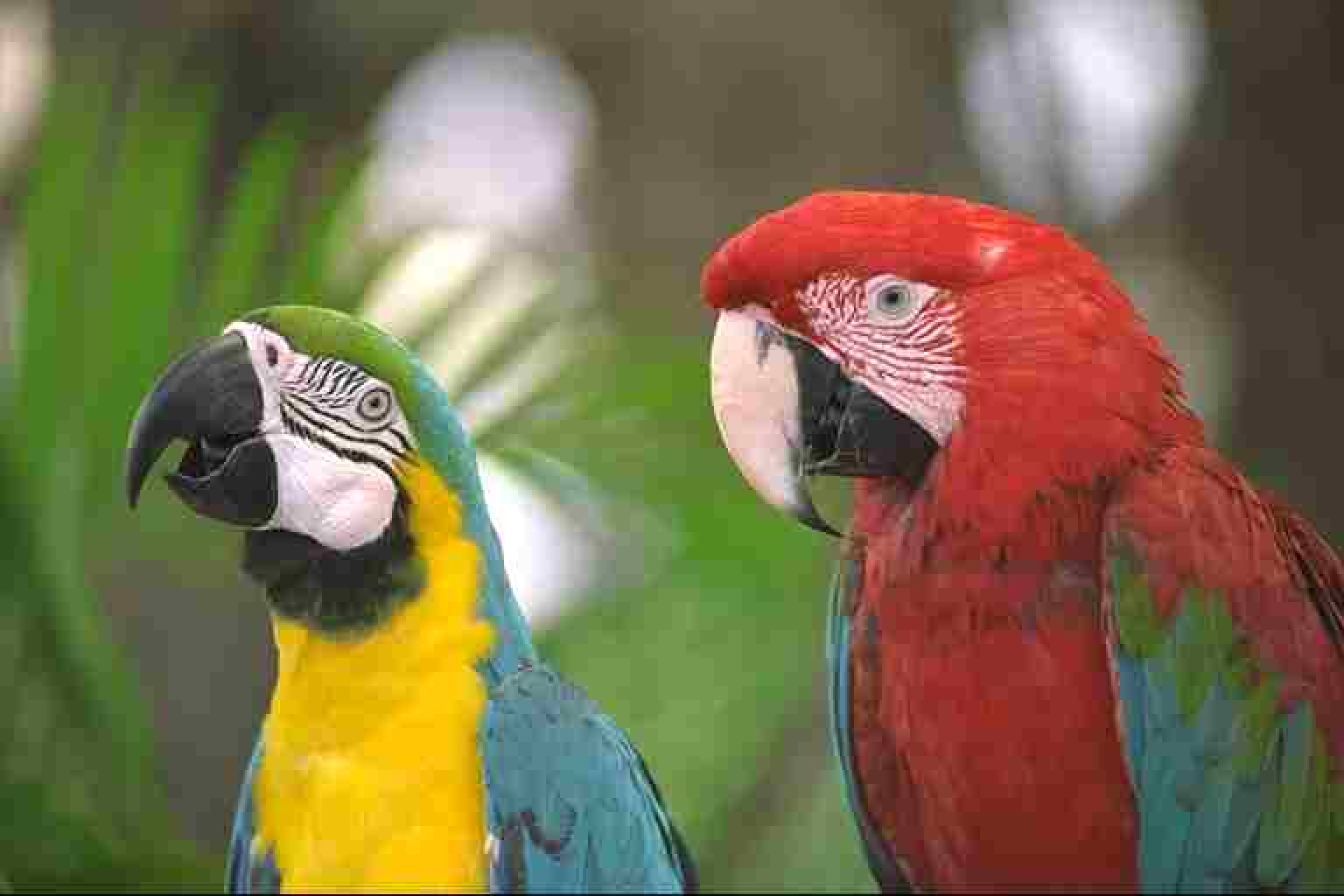}};
	\node at (12cm, 0cm) {\includegraphics[trim={1cm 2.5cm 8cm 2.5cm},clip,resolution=302]{figures/images/kodim23_0.25bpp_google.png}};

	\node at ( 0cm, -2cm) {\scriptsize\textsf{0.245972 bpp}};
	\node at ( 4cm, -2cm) {\scriptsize\textsf{0.250468 bpp}};
	\node at ( 8cm, -2cm) {\scriptsize\textsf{0.248413 bpp}};
	\node at (12cm, -2cm) {\scriptsize\textsf{0.25 bpp}};

	\node at ( 0cm, -4cm) {\includegraphics[trim={2.5cm 1.25cm 2.21cm 8cm},clip,resolution=302]{figures/images/kodim10_0.375bpp_cae.png}};
	\node at ( 4cm, -4cm) {\includegraphics[trim={2.5cm 1.25cm 2.21cm 8cm},clip,resolution=302]{figures/images/kodim10_0.375bpp_jp2.png}};
	\node at ( 8cm, -4cm) {\includegraphics[clip,resolution=302]{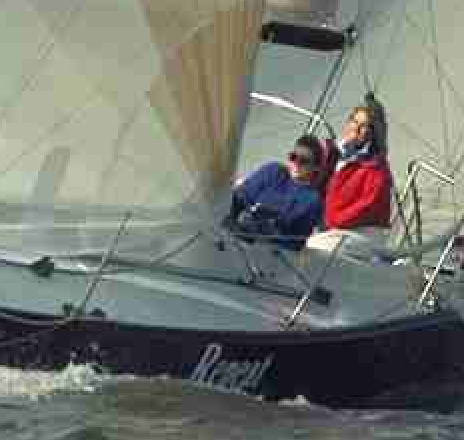}};
	\node at (12cm, -4cm) {\includegraphics[trim={2.5cm 1.25cm 2.21cm 8cm},clip,resolution=302]{figures/images/kodim10_0.375bpp_google.png}};

	\node at ( 0cm, -6cm) {\scriptsize\textsf{0.356608 bpp}};
	\node at ( 4cm, -6cm) {\scriptsize\textsf{0.359151 bpp}};
	\node at ( 8cm, -6cm) {\scriptsize\textsf{0.365438 bpp}};
	\node at (12cm, -6cm) {\scriptsize\textsf{0.375 bpp}};

	\node at ( 0cm, -8cm) {\includegraphics[trim={2.5cm 7.75cm 2.21cm 1.5cm},clip,resolution=302]{figures/images/kodim17_0.5bpp_cae.png}};
	\node at ( 4cm, -8cm) {\includegraphics[trim={2.5cm 7.75cm 2.21cm 1.5cm},clip,resolution=302]{figures/images/kodim17_0.5bpp_jp2.png}};
	\node at ( 8cm, -8cm) {\includegraphics[trim={2.5cm 7.75cm 2.21cm 1.5cm},clip,resolution=302]{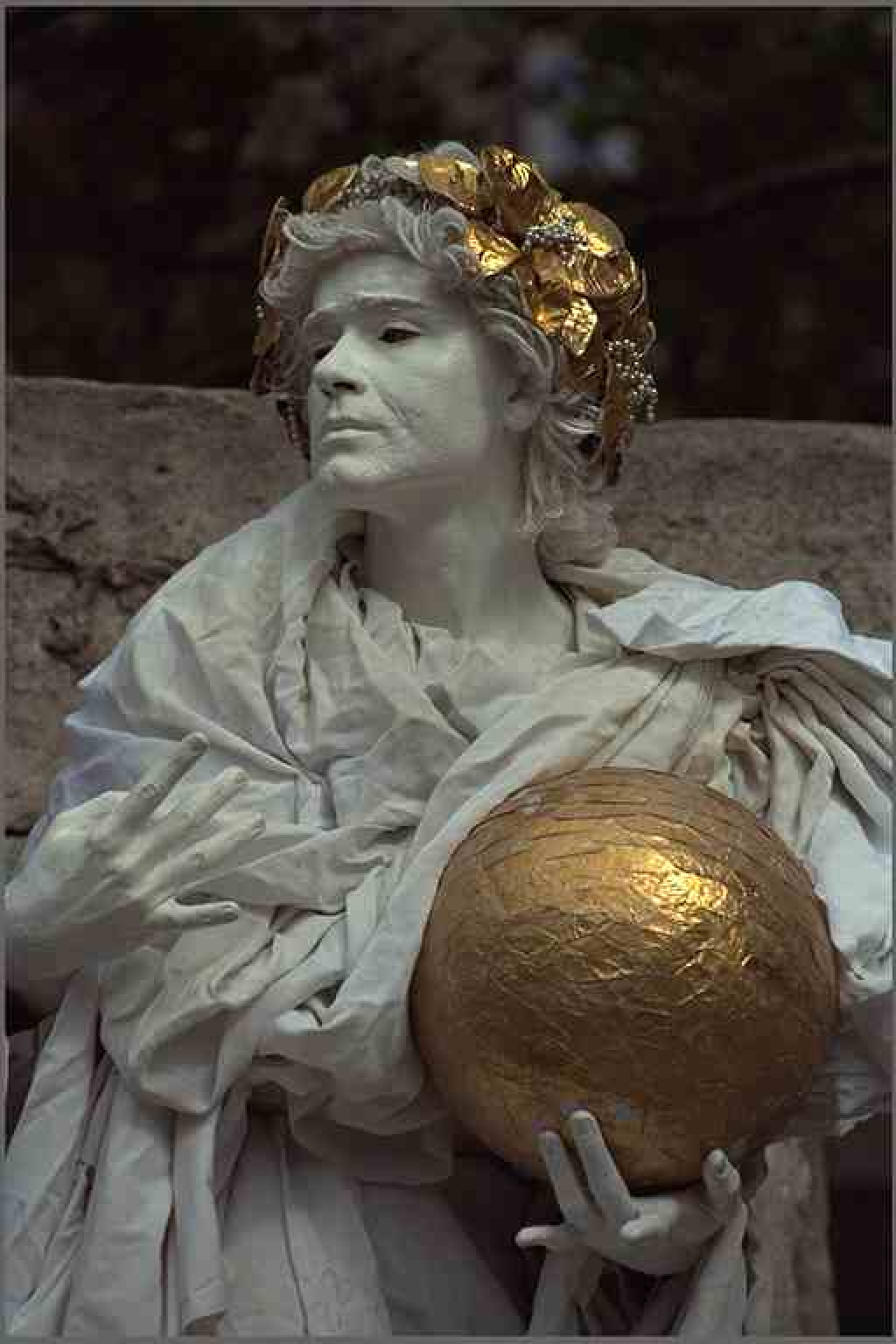}};
	\node at (12cm, -8cm) {\includegraphics[trim={2.5cm 7.75cm 2.21cm 1.5cm},clip,resolution=302]{figures/images/kodim17_0.5bpp_google.png}};

	\node at ( 0cm, -10cm) {\scriptsize\textsf{0.480632 bpp}};
	\node at ( 4cm, -10cm) {\scriptsize\textsf{0.491211 bpp}};
	\node at ( 8cm, -10cm) {\scriptsize\textsf{0.486755 bpp}};
	\node at (12cm, -10cm) {\scriptsize\textsf{0.5 bpp}};

	\node at ( 0cm, -12cm) {\includegraphics[trim={2.5cm 6.25cm 2.21cm 3cm},clip,resolution=302]{figures/images/kodim04_0.25bpp_cae.png}};
	\node at ( 4cm, -12cm) {\includegraphics[trim={2.5cm 6.25cm 2.21cm 3cm},clip,resolution=302]{figures/images/kodim04_0.25bpp_jp2.png}};
	\node at ( 8cm, -12cm) {\includegraphics[clip,resolution=302]{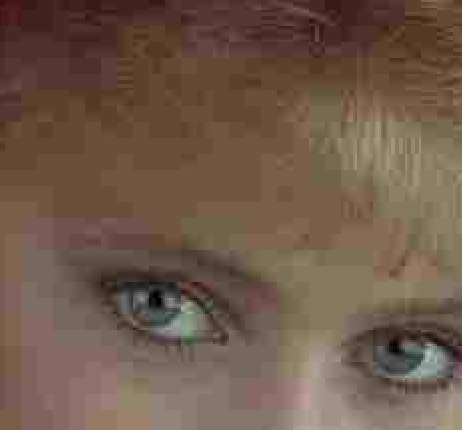}};
	\node at (12cm, -12cm) {\includegraphics[trim={2.5cm 6.25cm 2.21cm 3cm},clip,resolution=302]{figures/images/kodim04_0.25bpp_google.png}};

	\node at ( 0cm, -14cm) {\scriptsize\textsf{0.245626 bpp}};
	\node at ( 4cm, -14cm) {\scriptsize\textsf{0.249654 bpp}};
	\node at ( 8cm, -14cm) {\scriptsize\textsf{0.254415 bpp}};
	\node at (12cm, -14cm) {\scriptsize\textsf{0.25 bpp}};

	\node at ( 0cm, -16cm) {\includegraphics[trim={2.5cm 6.25cm 2.21cm 3cm},clip,resolution=302]{figures/images/kodim04_0.5bpp_cae.png}};
	\node at ( 4cm, -16cm) {\includegraphics[trim={2.5cm 6.25cm 2.21cm 3cm},clip,resolution=302]{figures/images/kodim04_0.5bpp_jp2.png}};
	\node at ( 8cm, -16cm) {\includegraphics[clip,resolution=302]{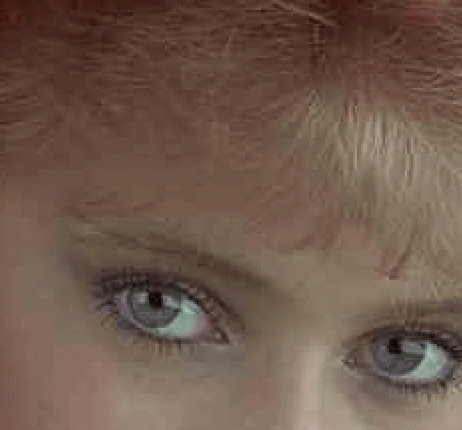}};
	\node at (12cm, -16cm) {\includegraphics[trim={2.5cm 6.25cm 2.21cm 3cm},clip,resolution=302]{figures/images/kodim04_0.5bpp_google.png}};

	\node at ( 0cm, -18cm) {\scriptsize\textsf{0.499308 bpp}};
	\node at ( 4cm, -18cm) {\scriptsize\textsf{0.504496 bpp}};
	\node at ( 8cm, -18cm) {\scriptsize\textsf{0.505473 bpp}};
	\node at (12cm, -18cm) {\scriptsize\textsf{0.5 bpp}};
\end{tikzpicture}

%% file: figures/jpeg.tex
\begin{tikzpicture}
	\node at (3.0cm, 2.5cm) {\includegraphics[width=15cm]{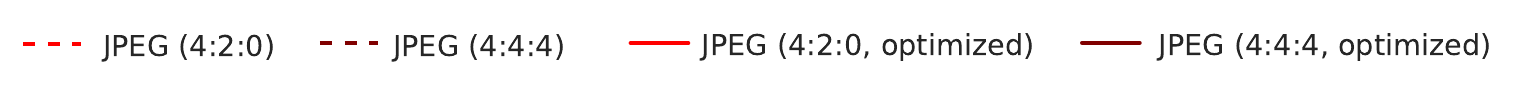}};

	\node[anchor=west] at (-4.7cm, 0cm) {\includegraphics[height=4.4cm]{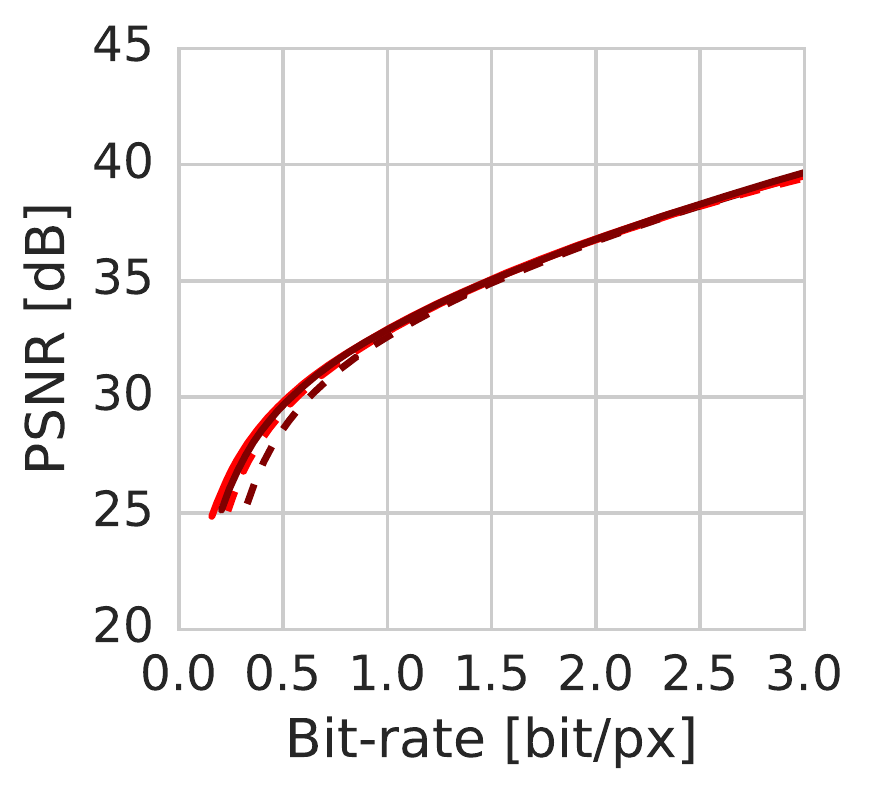}};
	\node[anchor=west] at ( 0.0cm, 0cm) {\includegraphics[height=4.4cm]{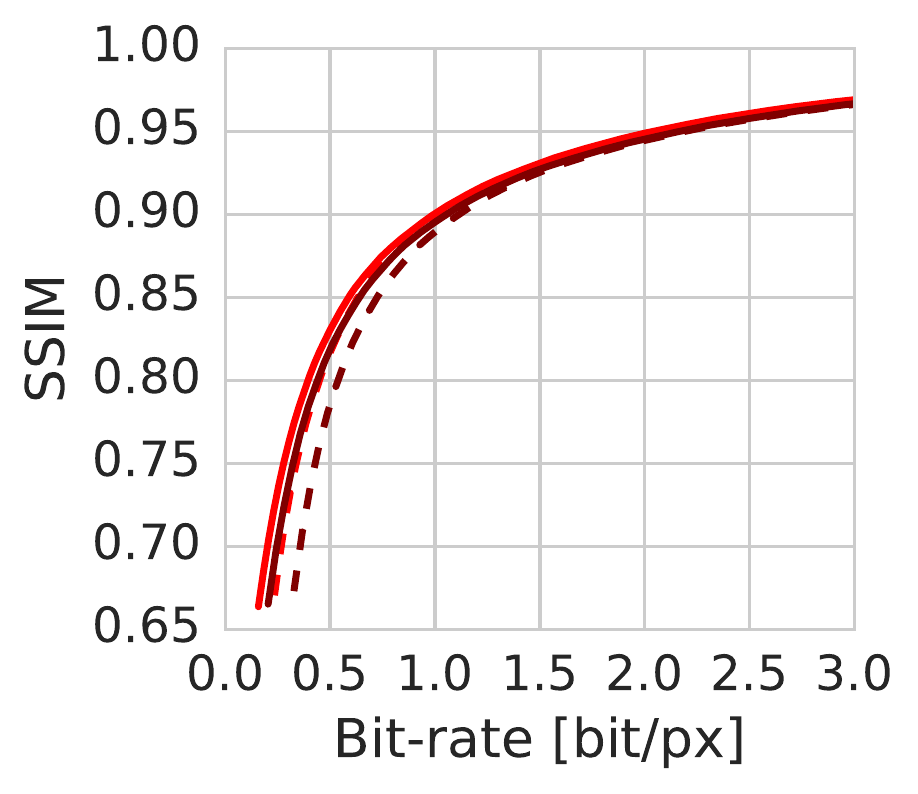}};
	\node[anchor=west] at ( 5.0cm, 0cm) {\includegraphics[height=4.4cm]{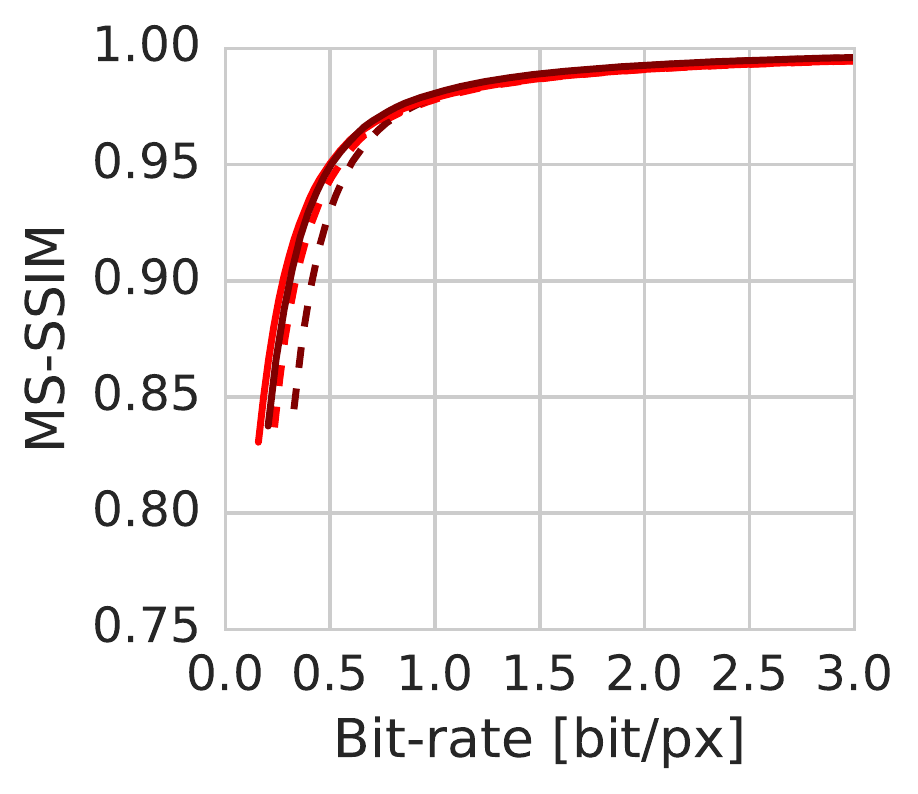}};
\end{tikzpicture}

%% file: figures/round_vs_noround.tex
\begin{tikzpicture}
	\node at (-4.9cm, 2.1cm) {\footnotesize\textsf{\textbf{A}}};
	\node at (-0.9cm, 2.1cm) {\footnotesize\textsf{\textbf{B}}};
	\node at ( 3.1cm, 2.1cm) {\footnotesize\textsf{\textbf{C}}};

	\node at (-3.2cm, 2.1cm) {\scriptsize\textsf{Original}};
	\node at ( 0.9cm, 2.1cm) {\scriptsize\textsf{Dim. reduction}};
	\node at ( 4.8cm, 2.1cm) {\scriptsize\textsf{Rounding}};

	\begin{scope}[xshift=.95cm]
		\node at ( -4cm, 0cm) {\includegraphics[clip,resolution=302]{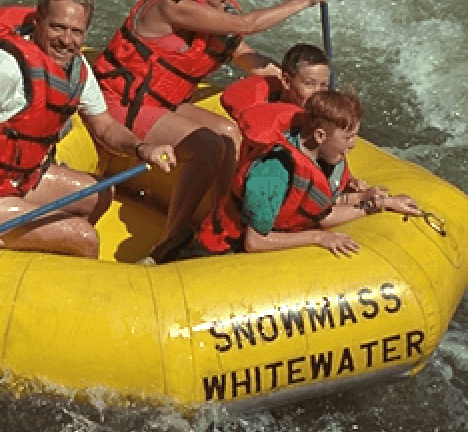}};
		\node at (  0cm, 0cm) {\includegraphics[clip,resolution=302]{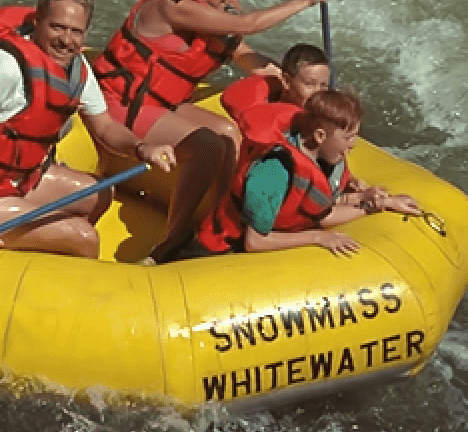}};
		\node at (  4cm, 0cm) {\includegraphics[clip,resolution=302]{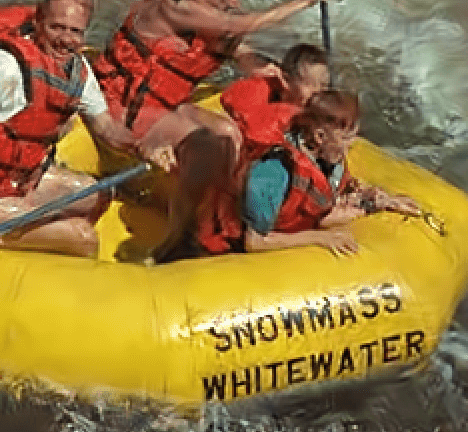}};
	\end{scope}
\end{tikzpicture}

%% file: figures/quantitative_comparison_cae.tex
\begin{tikzpicture}
	\node at (2.5cm, 2.5cm) {\includegraphics[width=11.5cm]{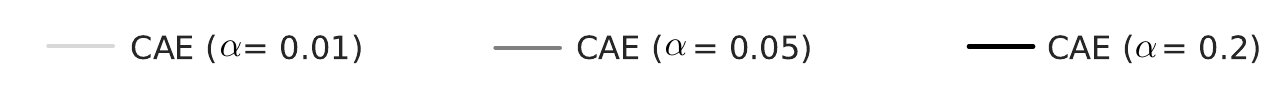}};

	\node[anchor=west] at (-4.7cm, 0cm) {\includegraphics[height=4.4cm]{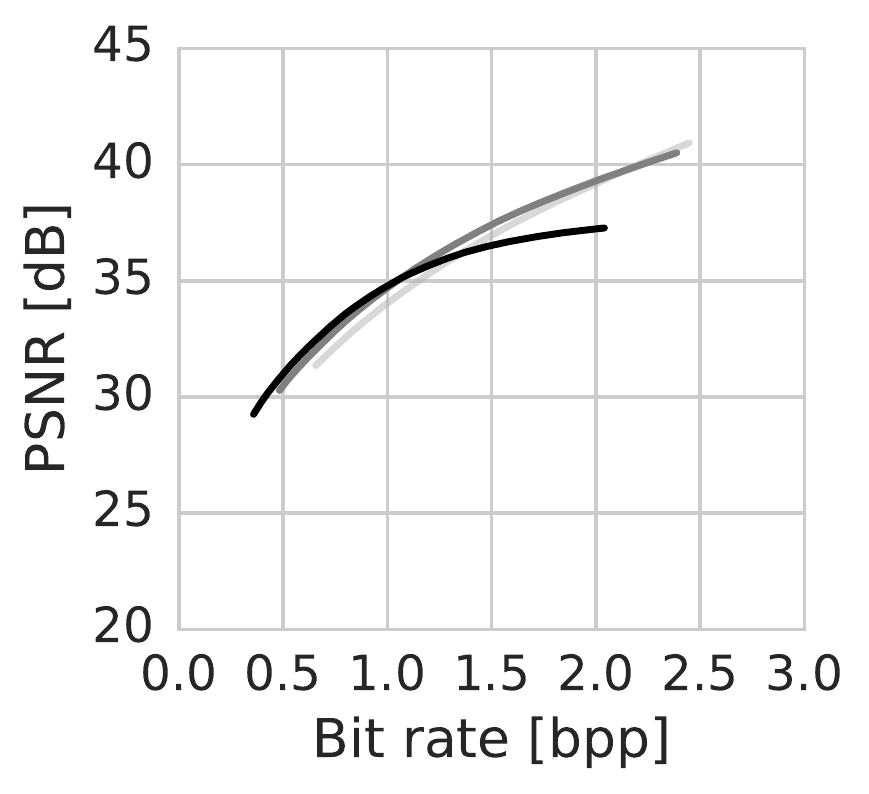}};
	\node[anchor=west] at ( 0.0cm, 0cm) {\includegraphics[height=4.4cm]{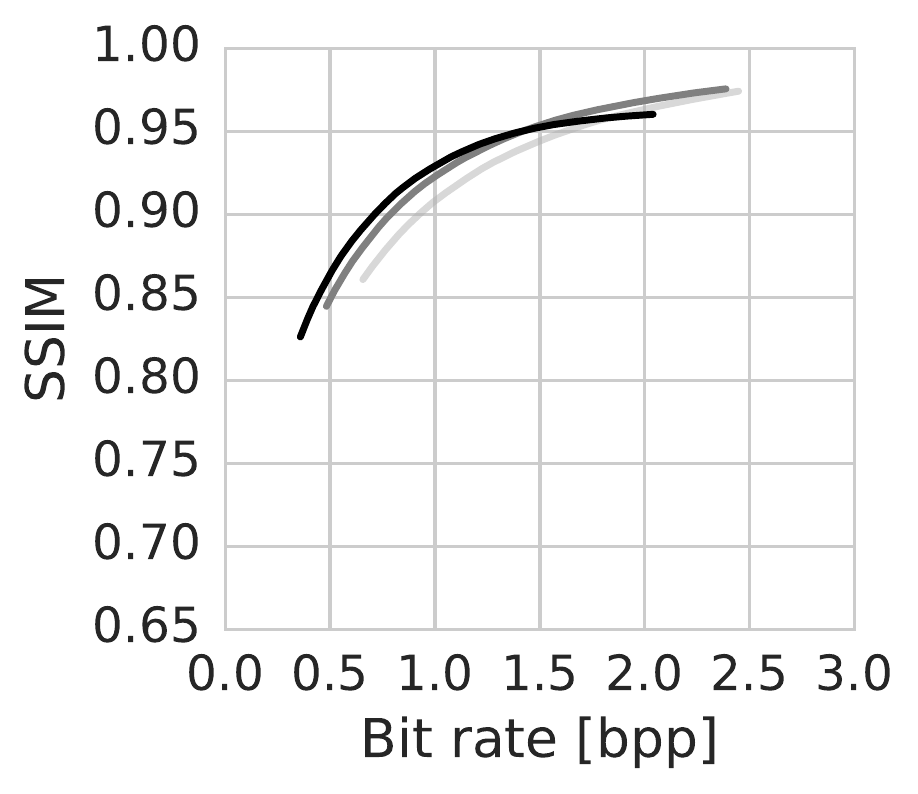}};
	\node[anchor=west] at ( 5.0cm, 0cm) {\includegraphics[height=4.4cm]{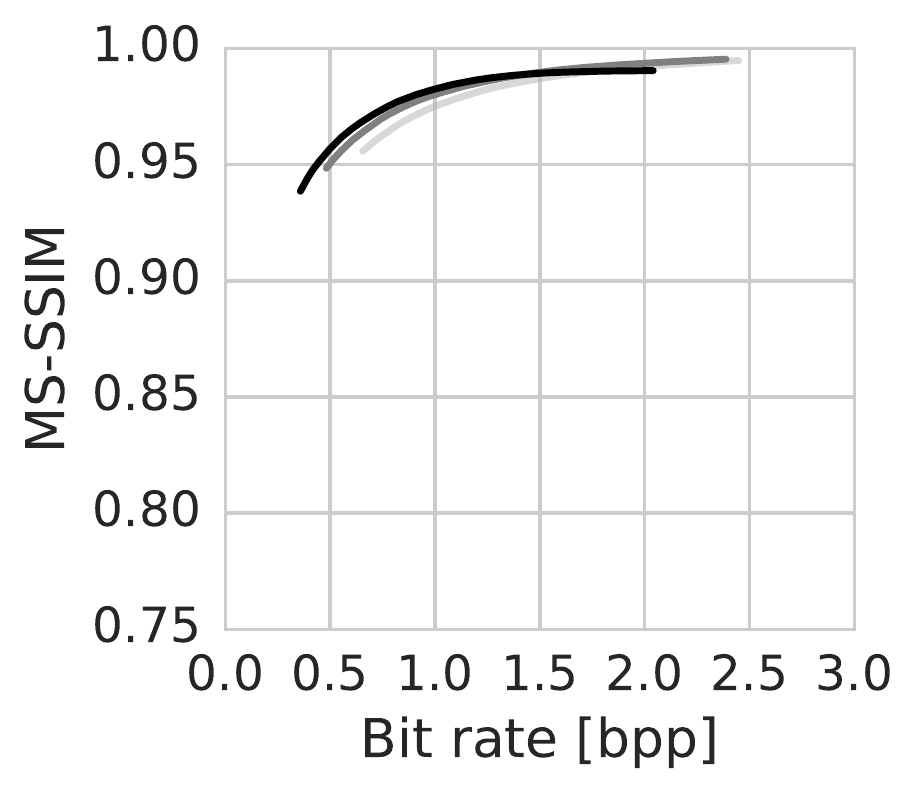}};
\end{tikzpicture}

%% file: figures/quantitative_comparison_balle.tex
\begin{tikzpicture}
	\node at (2.5cm, 2.5cm) {\includegraphics[width=5.2cm]{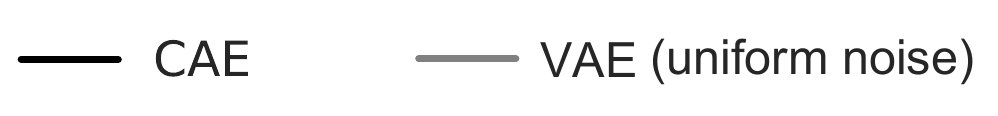}};

	\node[anchor=west] at (-4.7cm, 0cm) {\includegraphics[height=4.4cm]{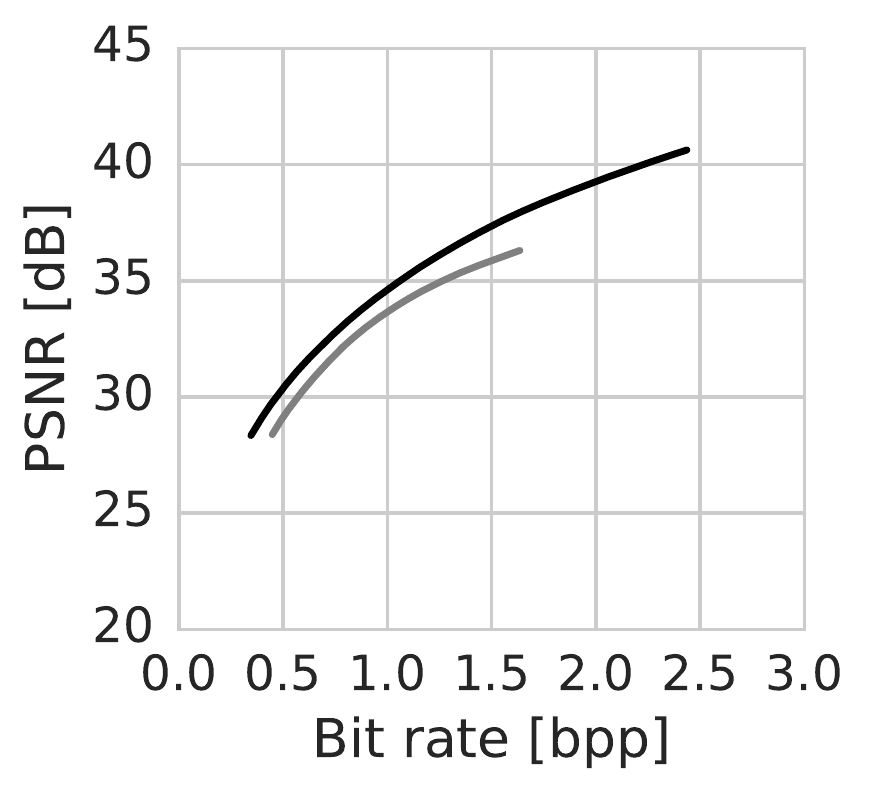}};
	\node[anchor=west] at ( 0.0cm, 0cm) {\includegraphics[height=4.4cm]{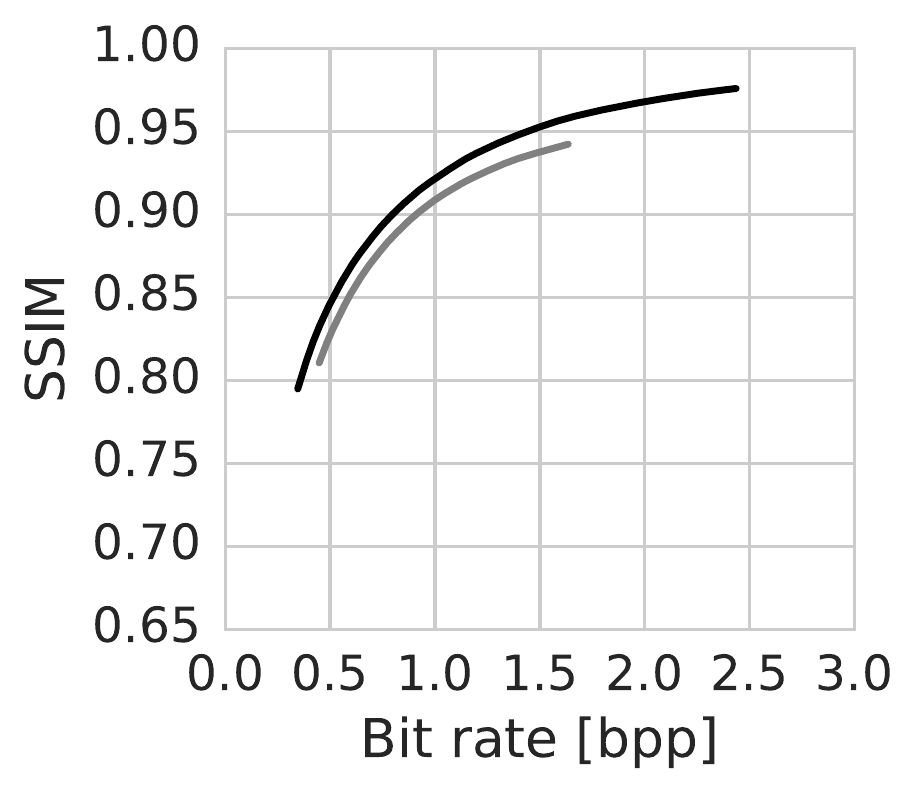}};
	\node[anchor=west] at ( 5.0cm, 0cm) {\includegraphics[height=4.4cm]{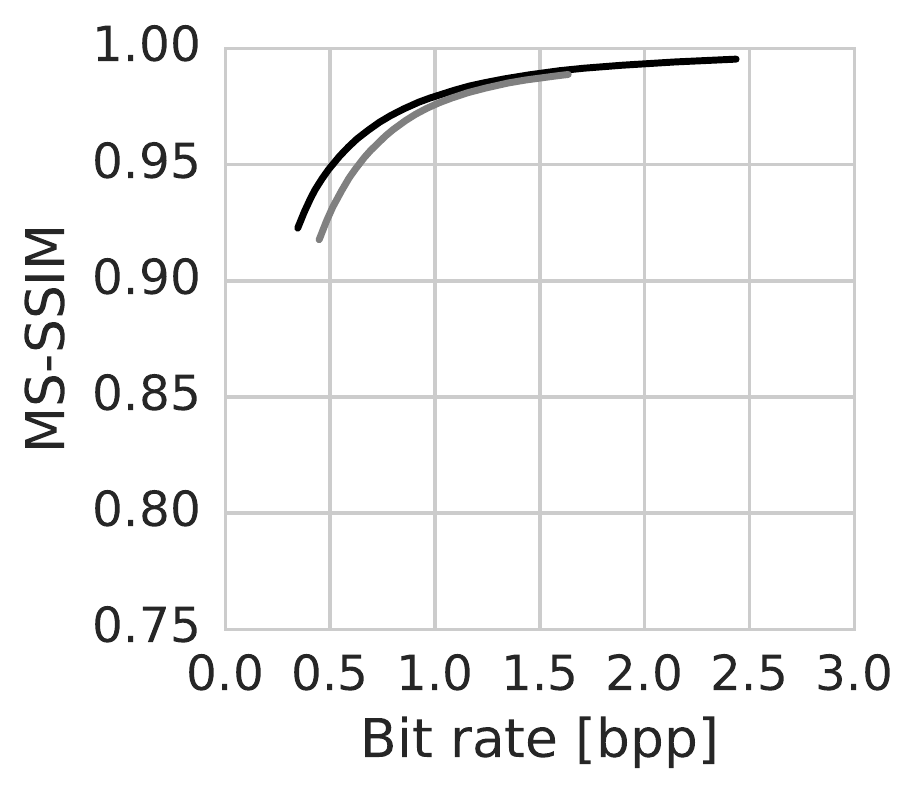}};
\end{tikzpicture}